%% file: iccv_final.tex
\documentclass[10pt,twocolumn,letterpaper]{article}

\usepackage{iccv}
\usepackage{times}
\usepackage{epsfig}
\usepackage{graphicx}
\usepackage{amsmath}
\usepackage{booktabs}
\usepackage{multirow}
\usepackage[T1]{fontenc}
\usepackage{amssymb}

\usepackage[pagebackref=true,breaklinks=true,letterpaper=true,colorlinks,bookmarks=false]{hyperref}

\iccvfinalcopy 

\def\iccvPaperID{3696} 

\ificcvfinal\fi

\begin{document}
\input{command}
\title{DRAW: Defending Camera-shooted RAW against Image Manipulation}

\author{Xiaoxiao Hu\textsuperscript{1,2*}, Qichao Ying\textsuperscript{1,2*}, Zhenxing Qian\textsuperscript{1,2$\dagger$}, Sheng Li\textsuperscript{1,2}, Xinpeng Zhang\textsuperscript{1,2}\\
\textsuperscript{1}School of Computer Science, Fudan University\\
\textsuperscript{2}Key Laboratory of Culture \& Tourism Intelligent Computing, Fudan University\\
}

\maketitle
\ificcvfinal\thispagestyle{empty}\fi

\newcommand\blfootnote[1]{%
\begingroup
\renewcommand\thefootnote{}\footnote{#1}%
\addtocounter{footnote}{-1}%
\endgroup
}

\blfootnote{*Xiaoxiao Hu and Qichao Ying contribute equally to this work.}
\blfootnote{$\dagger$Corresponding author: Zhenxing Qian (zxqian@fudan.edu.cn)}
\begin{abstract}
RAW files are the initial measurement of scene radiance widely used in most cameras, and the ubiquitously-used RGB images are converted from RAW data through Image Signal Processing (ISP) pipelines. Nowadays, digital images are risky of being nefariously manipulated. Inspired by the fact that innate immunity is the first line of body defense, we propose DRAW, a novel scheme of defending images against manipulation by protecting their sources, i.e., camera-shooted RAWs. Specifically, we design a lightweight Multi-frequency Partial Fusion Network (MPF-Net) friendly to devices with limited computing resources by frequency learning and partial feature fusion. It introduces invisible watermarks as protective signal into the RAW data. The protection capability can not only be transferred into the rendered RGB images regardless of the applied ISP pipeline, but also is resilient to post-processing operations such as blurring or compression. Once the image is manipulated, we can accurately identify the forged areas with a localization network. Extensive experiments on several famous RAW datasets, e.g., RAISE, FiveK and SIDD, indicate the effectiveness of our method. We hope that this technique can be used in future cameras as an option for image protection, which could effectively restrict image manipulation at the source.
\end{abstract}

\section{Introduction}

In the digital world, the credibility of the famous saying ``seeing is believing"  is largely at risk since nowadays people can easily manipulate critical content within an image and redistribute the fabricated version via the Internet.
Owing to the fact that readers are more susceptible to well-crafted misleading material, 
fabricated images can be a means for some politicians to sway public opinion.
In more severe cases, those fraudulent images can be used to bolster fake news or criminal investigation. 

Image manipulation detection~\cite{chen2008determining,popescu2005exposing} and localization~\cite{dong2021mvss, RIML} has become a critical area of research for decades, with the goal of 
distinguishing manipulated images from authentic ones and locating the manipulated areas.
While early methods mainly check the integrity of the images from statistical aspects, e.g., the Photo-Response Non-Uniformity (PRNU) noise~\cite{chen2008determining} and the fixed pattern noise (FPN)~\cite{kurosawa1999ccd},
the uprising of deep networks has greatly strengthened the capability to find traces left by a variety of manipulation~\cite{dong2021mvss,wu2019mantra,hu2020span}. 
\begin{figure}
  \includegraphics[width=0.48\textwidth]{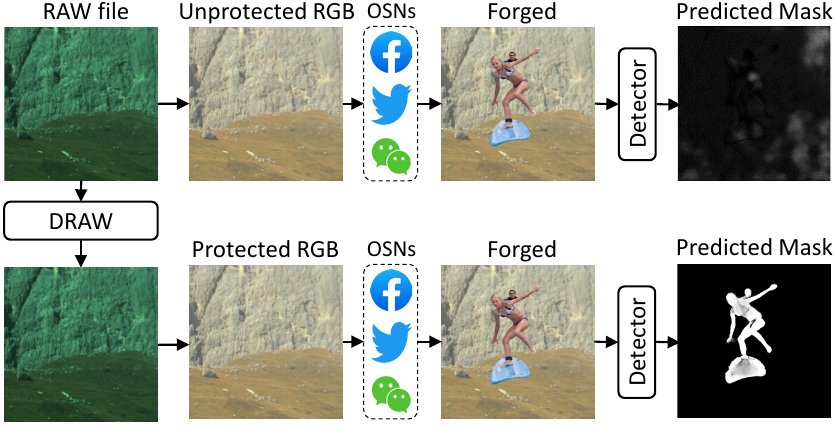}
  \caption{DRAW improves the performance of image manipulation localization against lossy image operations via imperceptible protective signal injection into RAW files.}
  \label{fig:teaser}
\end{figure}
However, the adversary is also continuously evolving both in strength and diversity.
For example, recent deep-network-based image editing algorithms~\cite{LAMA, ZITS} are reported to produce highly realistic images with almost no visible artifacts near the edges.
Therefore, it remains a big issue whether the learned subtle forensics traces can always be present in the newly forged images.
Also, though some works~\cite{RIML, wu2022robust} explicitly handle lossy online transmission scenarios,
they still face limited performance against well-crafted forgeries, e.g., inpainting, or lossy image operations, e.g., Gaussian blurring. 

Inspired by the fact that innate immunity is the first line of body defense and the best weapon to mitigate diseases, safeguarding images against manipulations is an alternative and promising way of deterring malicious attackers.
Indeed, the ubiquitous 8-bit RGB images are not the pristine format for reflecting how we perceive the world.
They are converted from RAW files via ISP pipelines.
Therefore, we propose DRAW, a proactive image protection scheme that defends camera-shooted RAW data against malicious manipulation on the RGB domain. 
Specifically, we propose to introduce imperceptible protective signal into the RAW data, which can be transferred into the rendered RGB images, even though various types of ISP pipelines are applied. 
Once these images are manipulated, the localization networks can exactly localize the forged areas regardless of image post-processing operations such as blurring, compression or color jittering.
Besides, a novel Multi-frequency Partial Fusion Network (MPF-Net) is proposed to implement RAW protection, which adopts frequency learning and cross-frequency partial feature fusion to significantly decrease the computational complexity.
We illustrate the functionality of DRAW in Fig.~\ref{fig:teaser}, which promotes accurate manipulation \greenmarker{localization} without affecting the visual quality.

Extensive experiments on several famous RAW datasets, e.g., RAISE, FiveK and SIDD, prove the imperceptibility, robustness and generalizability of our method.
Besides, to compare RAW-domain protection with previous works, we tempt to borrow the success of RGB-domain protection~\cite{asnani2022proactive,zhao2023proactive,ying2021image} as the baseline method for proactive manipulation localization.
The results show that DRAW hosts a noticeable performance gain and a nontrivial benefit of content-related adaptive embedding.
In addition, MPF-Net provides superior performance compared to classical U-Net~\cite{Unet} architecture with only 20.9\% of its memory cost and 0.95\% of its parameters. 
The novel lightweight architecture makes it possible to be integrated into cameras in the future, thereby changing the current situation where digital images can be freely manipulated.

The contributions of this paper are three-folded, namely:
\begin{enumerate}
\item DRAW is the first to propose
RAW protection against image manipulation. The corresponding RGB images will carry imperceptible protective signal even though various types of imaging pipelines or lossy image operations are applied. 
\item With RAW protection, image manipulation localization networks can better resist lossy image operations such as JPEG compression, blurring and rescaling.
\item  A novel lightweight MPF-Net is proposed for integrating RAW protection into cameras in the future, thereby potentially changing the current situation where digital images can be freely manipulated.
\end{enumerate}

\section{Related Works}
\noindent\textbf{Passive Image Manipulation Localization.}
Many existing image forensics schemes are designed to detect special kinds of attacks, e.g., splicing detection~\cite{RIML,salloum2018image}, copy-moving detection~\cite{islam2020doa,li2018fast} and inpainting detection~\cite{zhu2018deep,li2019localization}.
In addition, some universal tampering detection schemes~\cite{dong2021mvss,wu2019mantra,hu2020span} exploit universal noise artifacts left by manipulation. Mantra-Net~\cite{wu2019mantra} uses fully convolutional networks, Z-Pooling and long short-term memory cells for pixel-wise anomaly detection.
MVSS-Net~\cite{dong2021mvss} jointly exploits the noise view and the boundary artifact using multi-view feature learning and multi-scale supervision.
RGB-N~\cite{zhou2018learning} additionally utilizes auto-generated data augmentation for training. 
RIML~\cite{RIML} includes adversarial training, where the lossy Online Social Network (OSN) transmission is simulated by modeling noise from different sources. 
However, these works are still limited in generalization to well-crafted manipulations or heavy lossy operations. 
\begin{figure}
  \includegraphics[width=0.48\textwidth]{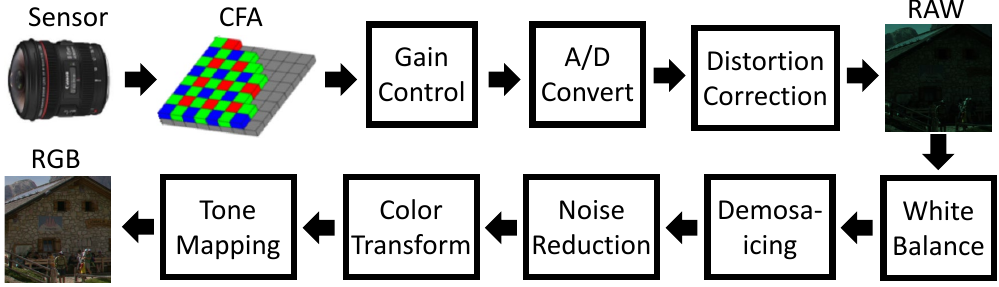}
  \caption{
  Typical camera imaging pipeline for RAW data acquisition and subsequent RGB image signal processing.}
  \label{demo_RAWtoRGB}
\end{figure}

\begin{figure*}[!t]
    \centering
    \includegraphics[width=1.0\textwidth]{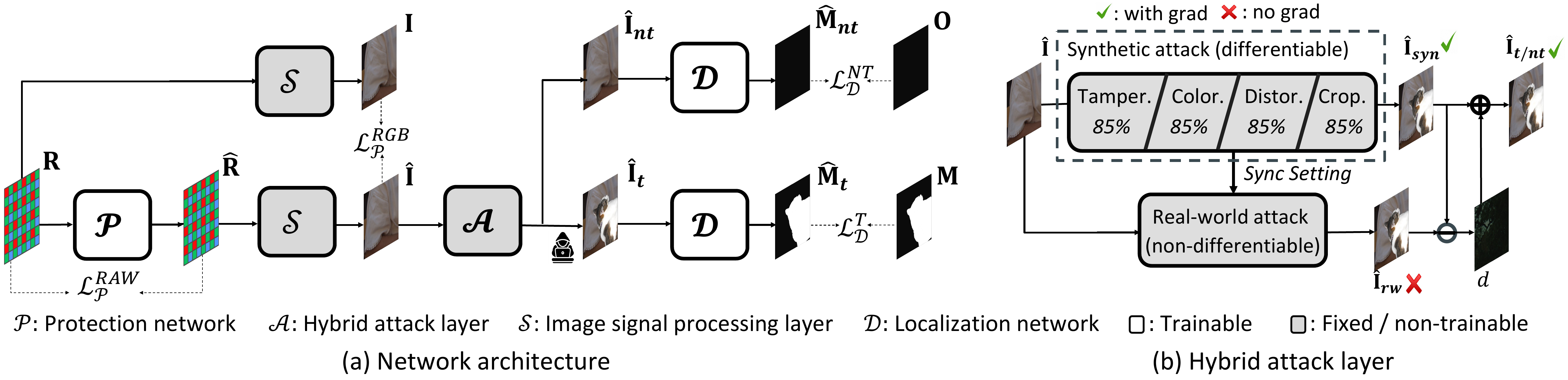}
    \caption{\textbf{Pipeline design of DRAW.} We design a lightweight protection network that embeds imperceptible protective signal in the RAW domain and transfers it into the rendered RGB images. On the recipient's side, the localization network identifies the forged areas.}
    \label{image_framework}
\end{figure*}
\noindent\textbf{Watermarking for Image Protection.}
Many image protection schemes based on watermarking~\cite{lu2021large,jing2021hinet,fridrich2012rich,ying2019robust} have been proposed.
Asnani et al.~\cite{asnani2022proactive} propose to embed templates into images for more accurate manipulation detection.
Zhao et al.~\cite{zhao2023proactive} embed watermarks as anti-Deepfake labels into the facial identity features.
FakeTagger~\cite{wang2021faketagger} embeds the identity information into the whole facial image, which can be recovered after illegal face swapping.
Khachaturov et al.~\cite{khachaturov2021markpainting} and Yin et al~\cite{yin2018deep} respectively propose to attack inpainting or Super-Resolution (SR) models by forcing them to work abnormally on the targeted images. 
\greenmarker{However, these approaches do not tackle the issue of forgery localization, and many of them cannot combat lossy image operations.}
We alternatively introduce imperceptible protective signal into RAW data and transfer it into RGB images to aid robust manipulation localization.

\noindent\textbf{Models for Limited Computing Resources.}
Classical network architectures for segmentation-based tasks, e.g., U-Net~\cite{Unet} or FPN~\cite{FPN}, usually require non-affordable computing resources for many small devices.
MobileNet~\cite{mobile} and ShuffleNet~\cite{ma2018shufflenet} are early works on addressing this issue respectively via Depth-wise Separable Convolution (DSConv) and channel split \& shuffle.
ENet~\cite{Enet} proposes an asymmetric encoder-decoder architecture with early downsampling.
SegNet~\cite{segnet} only stores the max-pooling indices of
the feature maps and uses them in its decoder network to
achieve good performance.
Despite substantial efforts made, these networks are either still computationally demanding or sacrifice performance for model size shrinkage.
We propose MPF-Net that contains only 20.9\% of memory cost and 0.95\% of parameters of U-Net yet provides surpassing performance in our task.

\section{Proposed Method}
\subsection{Approach}
Fig.~\ref{image_framework} depicts the pipeline design of DRAW. 
We denote the captured RAW data as $\mathbf{R}$, and use a protection network $\mathcal{P}$ to transform $\mathbf{R}$ into the protected RAW, i.e., $\hat{\mathbf{R}}$.
The functionality of $\mathcal{P}$ is to adaptively \ormarker{embed} a transferrable protective signal into $\hat{\mathbf{R}}$ for robust and accurate image manipulation localization in the RGB domain.
Considering the computational limitation of imaging equipment, we \ormarker{use a} novel lightweight MPF-Net specified in Section~\ref{section_network} \ormarker{to implement $\mathcal{P}$}. 
Next, we use the ISP layer $\mathcal{S}$ to render $\hat{\mathbf{R}}$ into the protected RGB image $\hat{\mathbf{I}}$.
Provided with a number of off-the-shelf deep-network-based ISP algorithms and non-differentiable conventional ISP algorithms, during training,
we include a popular conventional method, i.e., LibRaw~\cite{libraw} and two deep-learning methods, i.e.,
CycleISP~\cite{zamir2020cycleisp} and InvISP~\cite{InvISP}, and leave other ISP algorithms~\cite{zamir2022restormer,tradISP} for evaluation.
To improve generalizability,
interpolation is conducted on one network-rendered RGB $\hat{\mathbf{I}}_{\emph{net}}$ and one conventional-algorithm-generated RGB $\hat{\mathbf{I}}_{\emph{conv}}$ to produce $\hat{\mathbf{I}}$, i.e.,
$\hat{\mathbf{I}}=\omega\cdot\hat{\mathbf{I}}_{\emph{conv}}+(1-\omega)\cdot\hat{\mathbf{I}}_{\emph{net}}$,
where ${\omega}$ is uniformly within $[0,1]$. 

\begin{figure*}[!t]
    \centering
    \includegraphics[width=1.0\textwidth]{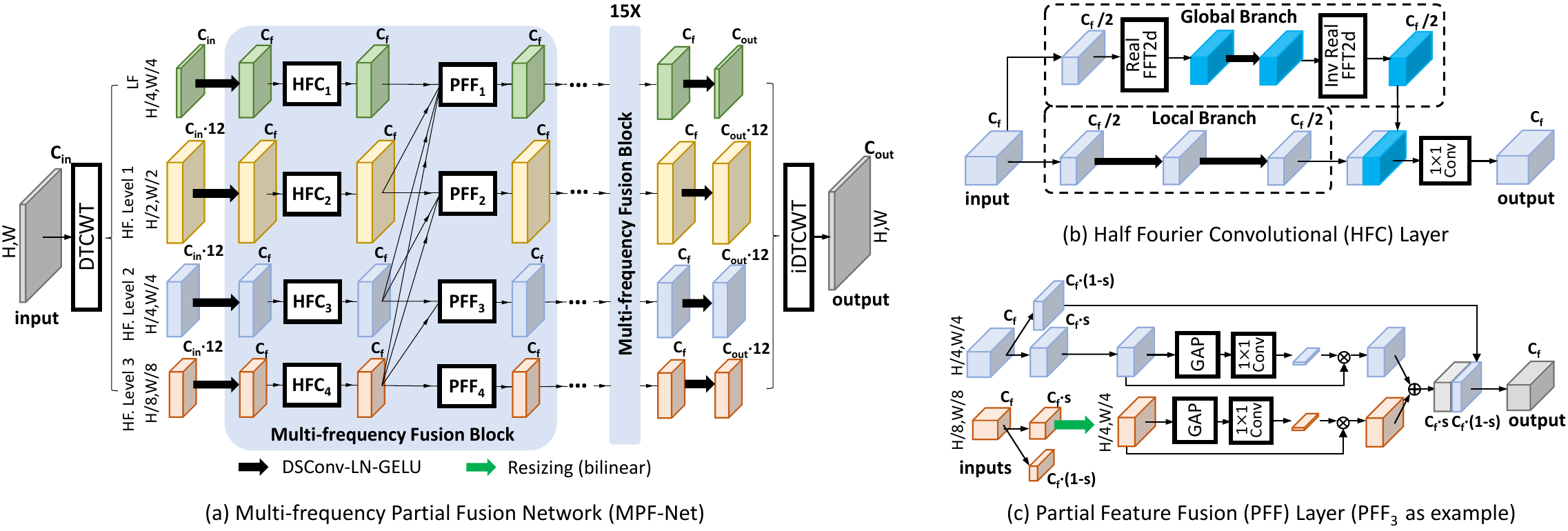}
    \caption{\textbf{Network design of Multi-frequency Partial fusion Network (MPF-Net).} It decomposes the input into multi-level subbands and during cross-frequency feature fusion, we preserve a proportion of features learned in the current layer. $C_\emph{in}=C_\emph{out}=3$ and $C_f=32$. }
    \label{image_hpn}
\end{figure*}
\begin{figure}[!t]
    \centering
    \includegraphics[width=0.48\textwidth]{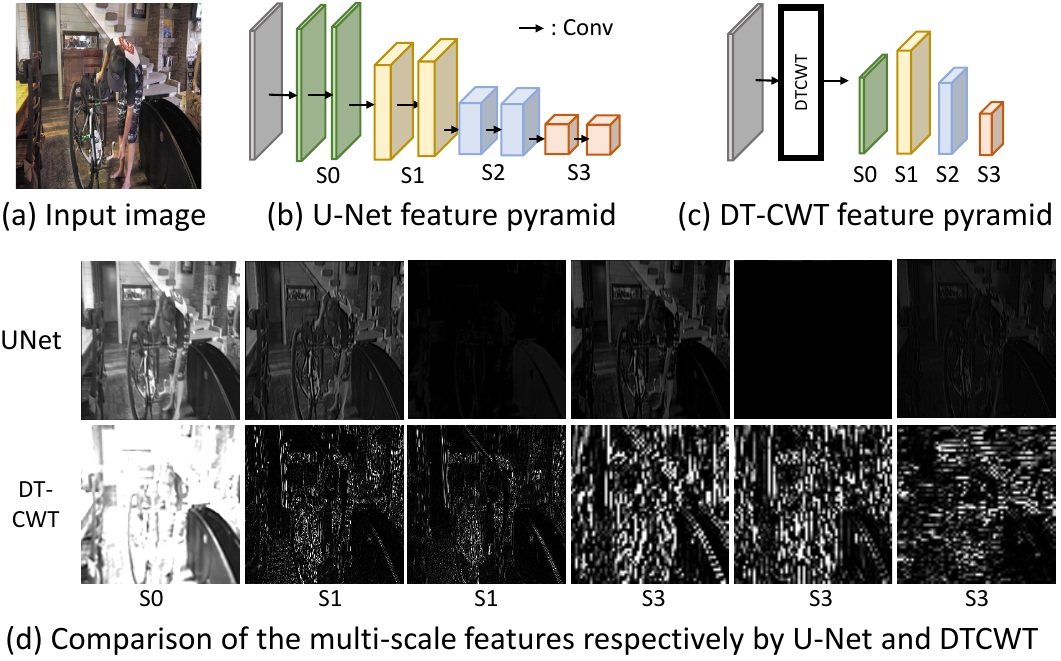}
    \caption{\textbf{Illustration of feature mining respectively using DT-CWT transform and U-Net.} DT-CWT requires fewer \textit{Conv} layers yet the generated features show less redundancy or repetition. }
    \label{DT-CWT_example}
\end{figure}
Afterward, to simulate image redistribution of $\hat{\mathbf{I}}$, we include the hybrid attack layer $\mathcal{A}$ to perform manipulation and lossy operations on $\hat{\mathbf{I}}$.
It comprises of modules for tampering, color adjustments, distortions (lossy operations) and cropping.
\greenmarker{
In line with typical forgery detection works~\cite{dong2021mvss,RIML}, we consider inpainting, splicing and copy-moving as the most common three types of tampering, which often alter the underlying meaning of an image.
In contrast, color adjustment and distortion are often considered benign yet can potentially erase traces for manipulation localization.
}
During training, these modules can be conditionally performed according to the empirical \textit{activation possibilities} (85\%) and in any arbitrary ordering to encourage diversity, e.g., tampering then distorting, cropping then tampering, etc.
We respectively denote the attacked images as $\hat{\mathbf{I}}_{\emph{t}}$ if the tampering module is activated or $\hat{\mathbf{I}}_{\emph{nt}}$ if otherwise. 
The latter is identified as authentic images, whose introduction is to explicitly minimize the false alarm rate of DRAW.
Detailed implementations of the modules are specified \redmarker{in the supplement}.
Besides, to closer the gap between real and simulated lossy operations and color jittering operations, 
we add the difference between $\hat{\mathbf{I}}_{\emph{syn}}$ and $\hat{\mathbf{I}}_{\emph{rw}}$ on to $\hat{\mathbf{I}}_{\emph{syn}}$, where
$\hat{\mathbf{I}}_{\emph{syn}}$ and $\hat{\mathbf{I}}_{\emph{rw}}$ respectively denote synthetic and real-world processed image using the same setting.
$x=\hat{\mathbf{I}}_{\emph{syn}}+\emph{sg}(\hat{\mathbf{I}}_{\emph{rw}}-\hat{\mathbf{I}}_{\emph{syn}}), x\in\{\hat{\mathbf{I}}_{\emph{t}},\hat{\mathbf{I}}_{\emph{nt}}\}$,
where $\emph{sg}$ stands for the stop-gradient operator~\cite{bengio2013estimating}.
 

On the recipient's side, we use the localization network $\mathcal{D}$ to estimate the manipulated region given a doubted image that could be one of $\hat{\mathbf{I}}_{\emph{t}}$ or $\hat{\mathbf{I}}_{\emph{nt}}$.
If it's an manipulated image $\hat{\mathbf{I}}_{\emph{t}}$, the predicted mask $\hat{\mathbf{M}}_t$ should be close to the ground-truth $\mathbf{M}$. 
Otherwise, it should be close to a zero matrix.
DRAW is flexible on the selection of $\mathcal{D}$, where many off-the-shelf networks can be applied, e.g., DRAW-HRNet~\cite{hrnet}, DRAW-MVSS~\cite{dong2021mvss} or DRAW-RIML~\cite{RIML}.
\begin{figure*}[!t]
    \centering
    \includegraphics[width=1.0\textwidth]{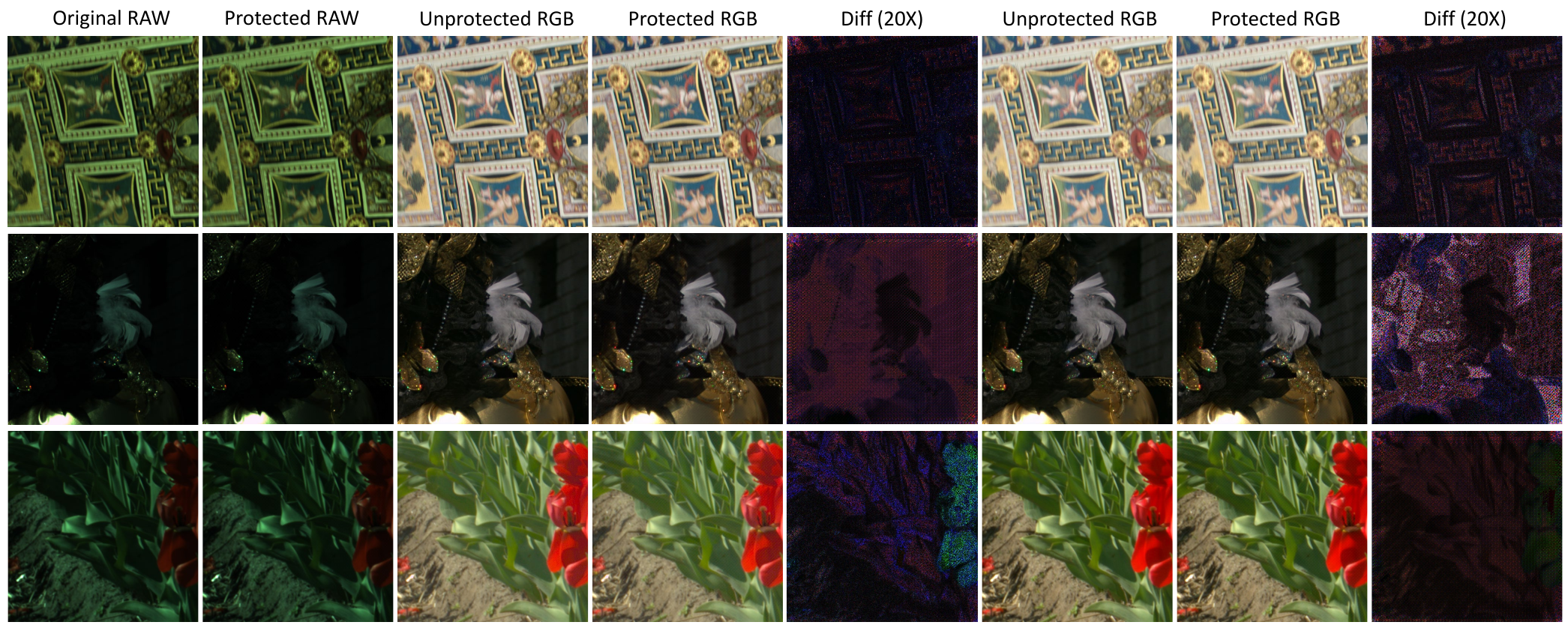}
    \caption{\textbf{Examples of protected images under different ISPs.} Dataset: RAISE. In each test, we apply two ISPs for rendering (upper: LibRAW / OpenISP, middle: InvISP / CycleISP, lower: OpenISP / InvISP). The RAW images are visualized through bilinear demosaicing.}
    \label{image_isp_samples}
\end{figure*}

\noindent\textbf{Objective Loss Functions.}
We need to include fidelity terms $\mathcal{L}_{\mathcal{P}}^{\emph{RAW}}$ and $\mathcal{L}_{\mathcal{P}}^{\emph{RGB}}$ to ensure imperceptible protection.
We find that the $\ell_1$ distance is the best in practice to minimize modification compared to many advanced deep-network-based terms, e.g., Lpips loss~\cite{zhang2018unreasonable} and contextual loss~\cite{zhang2019zoom}.
\begin{equation}
\label{loss_protect_RAW}
\begin{gathered}
\mathcal{L}_{\mathcal{P}}^{\emph{RAW}}=\mathbb{E}_{\mathbf{R}}\left[\left\|\mathbf{R}-\mathcal{P}\left(\mathbf{R}\right)\right\|_1\right], \\
\mathcal{L}_{\mathcal{P}}^{\emph{RGB}}=\mathbb{E}_{\mathbf{R}}\left[\left\|\mathcal{S}\left(\mathbf{R}\right)-\mathcal{S}\left(\mathcal{P}\left(\mathbf{R}\right)\right)\right\|_1\right].
\end{gathered}
\end{equation}
Next, we include localization terms
to minimize the Binary Cross Entropy (BCE) losses that respectively compare $\hat{\mathbf{M}}_t$ with $\mathbf{M}$, and $\hat{\mathbf{M}}_{\emph{nt}}$ with a zero matrix. 
\begin{equation}
\begin{gathered}
L_{\mathcal{D}}^{\emph{T}}= -\mathbb{E}_{\hat{\mathbf{I}}_{\emph{t}}}\left[\mathbf{M}\log \left(\mathcal{D}(\hat{\mathbf{I}}_{\emph{t}})\right)+\left(1-\mathbf{M}\right)\log\left(1-\mathcal{D}(\hat{\mathbf{I}}_{\emph{t}})\right)\right],\\
L_{\mathcal{D}}^{\emph{NT}}= -\mathbb{E}_{\hat{\mathbf{I}}_{\emph{nt}}}\left[\log\left(1-\mathcal{D}(\hat{\mathbf{I}}_{\emph{nt}})\right)\right].
\end{gathered}
\end{equation}
The total loss for DRAW is shown in Eq.~(\ref{eqn_loss_sum}), where $\alpha, \beta, \gamma, \epsilon$ are empirically-set hyper-parameters.
\begin{equation}
\begin{gathered}
\label{eqn_loss_sum}
\mathcal{L}=\alpha\cdot\mathcal{L}_{\mathcal{P}}^{\emph{RAW}}+\beta\cdot\mathcal{L}_{\mathcal{P}}^{\emph{RGB}}+\gamma\cdot\mathcal{L}_{\mathcal{D}}^{\emph{T}}+\epsilon\cdot\mathcal{L}_{\mathcal{D}}^{\emph{NT}},\\
\alpha=10, \beta=1,\gamma=0.02, \epsilon=0.01.
\end{gathered}
\end{equation}

\subsection{Multi-frequency Partial Fusion Network}
\label{section_network}
In order to combat sophisticated image manipulation within resource-limited environments such as cellphones and cameras, 
it is essential to deploy a lightweight architecture yet with rich feature extraction capabilities.
Fig.~\ref{image_hpn} illustrates the network design, where we first use a three-level DT-CWT transform to decompose the input into a low-frequency main component and three levels of higher-frequency subbands. 
Each level consists of six subbands in complex forms, representing different degrees of wavelet information. The real and imaginary parts of the subbands are then concatenated.
In Fig.~\ref{DT-CWT_example}, we compare the feature pyramid of U-Net to that of DT-CWT.
Vanilla convolutions can be less efficient due to the restriction of receptive field, feature redundancy, and repetition during training.
In contrast, DT-CWT provides a strong prior for mitigating these issues, requiring only one layer of separable convolution and yielding richer patterns within representations.

Following the initial feature extraction, 
we apply a ``DSConv-LN-GELU" layer to further refine the extracted features, which is in short for depth-wise separable convolution~\cite{mobile}, Layer Normalization~\cite{LN} and GELU activation~\cite{GELU}.
Next, we cascade sixteen multi-frequency partial fusion blocks in each level as feature refinement and fusion.
Each block contains a Half Fourier Convolution (HFC) layer and a Partial Feature Fusion (PFF) layer.
Notably, these blocks do not alter either the resolution or channel number of the features. 
Then we project the features back into the main components and three levels of subbands using another “DSConv-LN-GELU” layer, 
which are then transformed back into the RGB domain via iDT-CWT.


\noindent\textbf{Half Fourier Convolution Layer (HFC).}
We observe that features provided by DT-CWT
provide a rich local pattern, whereas the global information representation is lacking.
Considering that Fast Fourier Transform (FFT) is efficient in giving global information about the frequency components of an image~\cite{zhou2022fedformer,lee2021fnet},
we include both vanilla \textit{Conv} layer and Fast Fourier Transform (FFT) in each HFC to enable simultaneous global and local feature mining. For the HFC layer at level $i$:
\begin{equation}
\begin{gathered}
\emph{HFC}_{i}: \emph{output}=[\emph{GB}(\emph{input}_1), \emph{LB}(\emph{input}_2)],\\ \emph{input}=[\emph{input}_1,\emph{input}_2],
\end{gathered}
\end{equation}
where we evenly split the input tensor by half, send them respectively into the Global Branch (GB) and Local Branch (LB) of the HFC layer, and concatenate the resultant features. 
GB contains FFT, \textit{Conv} layer and inverse FFT.
LB is composed of a cascade of two vanilla \textit{Conv} layers.

\noindent\textbf{Partial Feature Fusion Layer (PFF).}
On fusing different groups of features, two most commonly-accepted ways are ``concatenate-and-reduce"~\cite{cho2021rethinking,hrnet} or ``attend-to-aggregate"~\cite{dual_attention,SKFF}.
We propose a novel paradigm of ``reserve-attend-and-assemble". Specifically, we split the input features into two halves based on a predetermined ratio $s$ (default 0.25), i.e., $\emph{input}_i=[\emph{input}_{i,1},\emph{input}_{i,2}]$ for PFF at level i.
The first half of the multi-level features ($C_f\cdot~s$) are resized into the size of the current level, and then separately reweighed using channel attention (CA).
Next, ``assemble" is done by pixel-wisely aggregating all groups of reweighed features and concatenating them with the reserved second half ($C_f\cdot~(1-s)$). 
Our paradigm can potentially mitigate the issue of over-attention on certain frequencies or covariance drift of the preserved representation, especially from shallow layers, caused by residual learning.
Furthermore, we only pass higher-frequency subbands into lower levels, which also encourages each level to process unique combinations of frequencies which reduces redundancy.
The operations in PFF at level $i$ can be mathematically defined as follows.
\begin{equation}
\begin{gathered}
\emph{PFF}_{i}: \emph{output}=[\emph{input}_{i,2}, \sum_{j\leq~i}\emph{CA}(\emph{Resize}(\emph{input}_{j,1}))]
\end{gathered}
\end{equation}
where $\emph{CA}$ is composed of a global average pooling layer and a $1\times1$ bottleneck convolution.



\section{Experiments}
\subsection{Experimental Setups}
We use RAISE~\cite{dang2015raise} dataset (8156 image pairs) and Canon subset (2997 image pairs) from the FiveK~\cite{bychkovsky2011learning} dataset as the training set.
Meanwhile, RAISE, Canon subset and Nikon subset (1600 image pairs) from FiveK as well as SIDD dataset~\cite{SIDD_2018_CVPR} are used to evaluate DRAW.
We divide them into training sets and test sets at a ratio of 85: 15.
We crop each RAW image into non-overlapping sub-images sized $512\times~512$. 
For quantitative analysis, inspired by ~\cite{zhou2018learning, RIML}, we opt to arbitrarily select regions for copy-moving and inpainting and borrow segmentation masks and the sources from MS-COCO~\cite{lin2014microsoft} dataset for splicing.
For qualitative analysis, we also manually manipulate over one hundred protected images and show some of the representative examples in the figures. 

We train our benchmark model by jointly training $\mathcal{P}$ with HRNet~\cite{hrnet} as $\mathcal{D}$. 
We then fix $\mathcal{P}$ and respectively training MVSS~\cite{dong2021mvss} and RIML~\cite{RIML} as $\mathcal{D}$ on top of the protected RGB images.
All models are trained with batch size 16 on four distributed NVIDIA RTX 3090 GPUs, and we train the networks for 10 epochs in roughly one day.
For gradient descent, we use Adam optimizer with the default hyper-parameters.
The learning rate is $1\times10^{-4}$.  
\begin{table}[!t]
\footnotesize
\renewcommand{\arraystretch}{1}
    \setlength{\tabcolsep}{1.5mm}
	\caption{\textbf{Quantitative analysis on the imperceptibility of RAW protection.} $[\mathbf{R},\hat{\mathbf{R}}]$: RAW file before and after protection. $[\mathbf{I},\hat{\mathbf{I}}]$: RGB file rendered respectively from $\mathbf{R}$ and $\hat{\mathbf{R}}$ using different ISP pipelines. Dataset: RAISE and Canon.}
	\label{table_different_resolution_protected}
	\centering
	\begin{tabular}{c|cc|cc|cc}
		\hline
		\multirow{2}{*}{Process} & \multicolumn{2}{c|}{$512\times512$} & \multicolumn{2}{c|}{$256\times256$} & \multicolumn{2}{c}{$1024\times1024$} \\
		& PSNR & SSIM & PSNR & SSIM & PSNR & SSIM
		\\
        \hline
        [$\mathbf{R}$,$\hat{\mathbf{R}}$] & 58.43 & - & 61.67 & - & 56.41 & - \\
    
        [$\mathbf{I}$,$\hat{\mathbf{I}}$] (InvISP) & 45.13 & 0.977 & 46.20 & 0.985 & 45.60 & 0.983 \\
    
	
        [$\mathbf{I}$,$\hat{\mathbf{I}}$] (LibRaw) & 41.25 & 0.960 & 41.97 & 0.967 & 41.07 & 0.957 \\
        
        [$\mathbf{I}$,$\hat{\mathbf{I}}$] (Restormer) & 45.75 & 0.980 & 46.24 & 0.984 & 45.03 & 0.977 \\
        
        [$\mathbf{I}$,$\hat{\mathbf{I}}$] (OpenISP) & 40.52 & 0.960 & 41.95 & 0.966 & 40.34 & 0.955 \\
		\hline
	\end{tabular}
\end{table}
\setlength{\tabcolsep}{1.1mm}{
\begin{table*}
\footnotesize
\renewcommand{\arraystretch}{1}
    \caption{\textbf{Average performance of different methods on forgery localization.} Dataset: RAISE. The best performances are highlighted in bold type. *: open-source pretrained models finetuned on original RAISE images with \textit{copy-moving}, \textit{splicing} and \textit{inpainting}.}
    	\label{table_comparison}
    \centering
    
	\begin{tabular}{c|c|ccc|ccc|ccc|ccc|ccc|ccc|ccc}
		\hline
		 &
		\multirow{2}{*}{Models} & 
		 \multicolumn{3}{c}{No attack} &
		 \multicolumn{3}{c}{Rescaling} &
		 \multicolumn{3}{c}{AWGN} & \multicolumn{3}{c}{JPEG90} & \multicolumn{3}{c}{JPEG70} & \multicolumn{3}{c}{Med. Blur} & \multicolumn{3}{c}{GBlur} \\
        \cline{3-23}
        & & Rec. & F1 & IoU 
        & Rec. & F1 & IoU 
        & Rec. & F1 & IoU 
        & Rec. & F1 & IoU 
        & Rec. & F1 & IoU  
        & Rec. & F1 & IoU 
        & Rec. & F1 & IoU \\
        \hline
        \multirow{5}{*}{\rotatebox[origin=c]{90}{\textit{splicing}}} 
        & MVSS${^{*}}$
        & .908 & .725 & .597 & .715 & .609 & .470 & \textbf{.954} & .688 & .547 & \textbf{.944} & .627 & .481 & \textbf{.915} & .565 & .415 & .869 & .695 & .561 & .181 & .211 & .138  \\
        & RIML${^{*}}$
        & \textbf{.941} & \textbf{.949} & \textbf{.908} & .732 & .795 & .702 & .900 & .918 & .863 & .869 & .892 & .821 & .777 & .818 & .721 & .900 & .918 & .857 & .096 & .142 & .094  \\
        & DRAW-MVSS 
        & .867 & .874 & .793 & .553 & .636 & .514 & .886 & .854 & .764 & .878 & .856 & .767 & .820 & .789 & .680 & .732 & .770 & .658 & .320 & .419 & .301 \\
        & DRAW-RIML 
        & .897 & .926 & .876 & .877 & .910 & .856 & .928 & \textbf{.946} & \textbf{.905} & .913 & .932 & .884 & .889 & \textbf{.909} & \textbf{.849} & .917 & .939 & \textbf{.893} & \textbf{.556} & \textbf{.639} & \textbf{.544}  \\
        & DRAW-HRNet 
        & .936 & .947 & .903 & \textbf{.922} & \textbf{.934} & \textbf{.884} & .929 & .934 & .883 & .933 & \textbf{.935} & \textbf{.885} & .902 & .861 & .776 & \textbf{.927} & \textbf{.940} & .891 & .552 & .638 & .523  \\
		\hline
		\multirow{5}{*}{\rotatebox[origin= c]{90}{\textit{copy-moving}}} 
        & MVSS${^{*}}$
        & .833 & .781 & .703 & .677 & .636 & .544 & .861 & .755 & .668 & .771 & .627 & .527 & .653 & .471 & .366 & .795 & .731 & .640 & .339 & .336 & .258 \\
        & RIML${^{*}}$  
        & .888 & .889 & .856 & .774 & .793 & .737 & .896 & .895 & .861 & .829 & .835 & .788 & .694 & .719 & .657 & .850 & .856 & .811 & .557 & .572 & .493 \\
        & DRAW-MVSS 
        & .901 & .893 & .857 & .839 & .836 & .780 & .915 & .890 & .850 & .862 & .842 & .793 & .804 & .767 & .706 & .871 & .851 & .803 & .631 & .657 & .582 \\
        & DRAW-RIML 
        & .915 & .925 & .910 & .875 & .895 & .868 & .906 & .918 & .899 & .884 & .899 & .874 & .845 & .866 & .829 & .897 & .910 & .888 & .774 & .811 & .768  \\
        & DRAW-HRNet 
        & \textbf{.969} & \textbf{.970} & \textbf{.959} & \textbf{.960} & \textbf{.956} & \textbf{.937} & \textbf{.962} & \textbf{.957} & \textbf{.943} & \textbf{.955} & \textbf{.951} & \textbf{.932} & \textbf{.916} & \textbf{.884} & \textbf{.839} & \textbf{.958} & \textbf{.955} & \textbf{.939} & \textbf{.915} & \textbf{.920} & \textbf{.885}  \\
		\hline
		\multirow{5}{*}{\rotatebox[origin= c]{90}{\textit{inpainting}}} 
        & MVSS${^{*}}$
        & .259 & .229 & .172 & .101 & .062 & .039 & .404 & .360 & .263 & .180 & .090 & .054 & .212 & .097 & .058 & .088 & .050 & .030 & .085 & .043 & .026  \\
        & RIML${^{*}}$
        & .126 & .140 & .097 & .035 & .047 & .030 & .132 & .155 & .113 & .014 & .020 & .013 & .001 & .001 & .001 & .037 & .043 & .026 & .068 & .077 & .048  \\
        & DRAW-MVSS 
       & .737 & .752 & .672 & .657 & .682 & .588 & .771 & .756 & .667 & .617 & .645 & .546 & \textbf{.515} & \textbf{.536} & \textbf{.434} & .567 & .595 & .497 & .514 & .561 & .463   \\
        & DRAW-RIML 
        & .663 & .716 & .656 & .457 & .518 & .452 & .667 & .718 & .654 & .348 & .411 & .342 & .091 & .121 & .089 & .366 & .423 & .360 & .284 & .338 & .281 \\
        & DRAW-HRNet 
        & \textbf{.776} & \textbf{.791} & \textbf{.735} & \textbf{.754} & \textbf{.760} & \textbf{.685} & \textbf{.788} & \textbf{.771} & \textbf{.697} & \textbf{.719} & \textbf{.714} & \textbf{.625} & .468 & .454 & .346 & \textbf{.732} & \textbf{.735} & \textbf{.647} & \textbf{.686} & \textbf{.704} & \textbf{.618}  \\
        \hline
	\end{tabular}
\end{table*}
}



\subsection{Performances}
\label{section_performance}
\noindent\textbf{Image Quality Assessment.}
Fig.~\ref{image_isp_samples} and Table~\ref{table_different_resolution_protected} respectively show the qualitative and quantitative results on the imperceptibility of the protection.
Besides, we test the overall image quality of protected images using untrained ISP network, namely, Restormer~\cite{zamir2022restormer}, and another conventional ISP, namely, OpenISP~\cite{tradISP}.
Restormer is originally proposed for image restoration, but we find that the transformer-based architecture also shows excellent performance on RGB image rendering.
OpenISP is another popular open-source ISP pipeline apart from LibRaw, and we customize the pipeline by applying the most essential modules.
We can observe little artifact from the protected version of RAW data and RGB. 
From the augmented difference, DRAW imperceptibly introduce content-related local patterns, which function like digital \textit{locks} onto the pixels and forgery localization is conducted by observing the integrity of these \textit{locks}.

\begin{figure}[!t]
    \centering
    \includegraphics[width=0.48\textwidth]{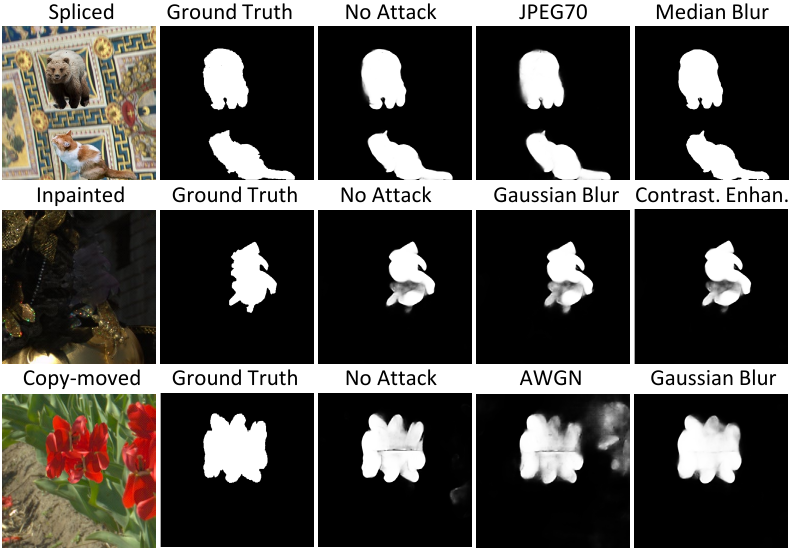}
    \caption{\textbf{Exampled forgery localization results of DRAW-HRNet.} The corresponding protected images are in Fig.~\ref{image_isp_samples}.}
    \label{image_detection_hrnet}
\end{figure}
\begin{figure*}[!t]
    \centering
\includegraphics[width=1.0\textwidth]{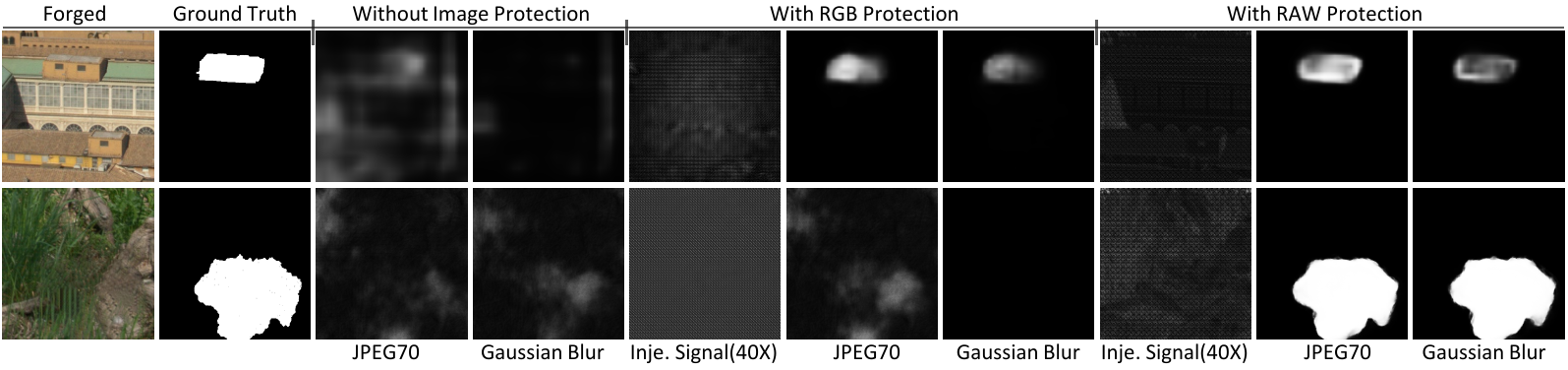}
    \caption{\textbf{Qualitative analysis on performance between passive localization without image protection, with RGB protection and with RAW protection.} Dataset: RAISE. $\mathcal{D}$: MVSS${^*}$ (upper), RIML${^*}$ (lower). Type: copy-moving (upper), inpainting (lower).}
    \label{image_comparison}
\end{figure*}

\noindent\textbf{Robustness and Accuracy of Manipulation Localization.}
We conduct comprehensive experiments on RAISE and Canon datasets under different lossy operations.
The qualitative and quantitative comparisons in terms of the Recall, F1 and IoU in the pixel domain are reported in Fig.~\ref{image_detection_hrnet}, Fig.~\ref{image_comparison}, Table~\ref{table_comparison} and Table~\ref{table_comparison_CANON}. 
The results under image color adjustment operations and combined attacks are included \redmarker{in the supplement}.
We find that for DRAW-HRNet, although the images are manipulated by diverse lossy operations, we succeed in localizing the tampered areas. 
If there are no lossy operations, the F1 scores are in most cases above 0.8.
Fig.~\ref{image_detection_hrnet} further provides exampled image manipulation localization results of DRAW-HRNet under different lossy operations.

Next, for fair comparison with previous arts, we finetune MVSS and RIML on RAISE and Canon dataset using the mechanisms proposed in the corresponding papers yet additionally considering \textit{splicing}, \textit{copy-moving} and \textit{inpainting}.
When heavy image lossy operations are present, MVSS fails to detect the tampered content. While RIML exhibits better robustness due to OSN transmission simulation, its performances under blurring or inpainting attacks are still restricted. 
However, training these detectors based on the protected images significantly improves their robustness.

\setlength{\tabcolsep}{1.1mm}{
\begin{table}
\footnotesize
\renewcommand{\arraystretch}{1}
    \caption{\textbf{Average performance of different methods on forgery localization.} Dataset: CANON.}
    	\label{table_comparison_CANON}
    \centering
    
	\begin{tabular}{c|c|cc|cc|cc|cc}
		\hline
		 &
		\multirow{2}{*}{Models} & 
		 \multicolumn{2}{c}{No attack} &
		 \multicolumn{2}{c}{Rescaling} &
            \multicolumn{2}{c}{JPEG70} &  \multicolumn{2}{c}{GBlur} \\
        \cline{3-10}
        & & F1 & IoU 
        & F1 & IoU 
        & F1 & IoU 
        & F1 & IoU \\
        \hline
        \multirow{5}{*}{\rotatebox[origin=c]{90}{\textit{splicing}}} 
        & MVSS${^{*}}$
        & .610 & .465 & .530 & .390 & .503 & .354 & .210 & .135 \\
        & RIML${^{*}}$
        & .925 & \textbf{.872} & .716 & .609 & .783 & .675 & .136 & .094 \\
        & DRAW-MVSS 
        & .841 & .738 & .875 & .789 & .842 & .739 & .829 & .731  \\
        & DRAW-RIML 
        & .887 & .818 & .925 & .870 & .855 & .769 & .906 & .843 \\
        & DRAW-HRNet 
        & \textbf{.926} & .869 & \textbf{.939} & \textbf{.889} & \textbf{.921} & \textbf{.861} & \textbf{.939} & \textbf{.891}  \\
		\hline
		\multirow{5}{*}{\rotatebox[origin= c]{90}{\textit{copy-moving}}} 
        & MVSS${^{*}}$
        & .727 & .628 & .624 & .526 & .455 & .341 & .371 & .283\\
        & RIML${^{*}}$  
        & .892 & .852 & .789 & .733 & .702 & .619 & .569 & .494  \\
        & DRAW-MVSS 
        & .912 & .879 & .868 & .828 & .832 & .785 & .826 & .776 \\
        & DRAW-RIML 
        & \textbf{.969} & \textbf{.957} & \textbf{.962} & \textbf{.947} & .911 & .882 & .952 & .933\\
        & DRAW-HRNet 
        & .968 & .956 & .960 & .945 & \textbf{.922} & \textbf{.895} & \textbf{.952} & \textbf{.934} \\
		\hline
		\multirow{5}{*}{\rotatebox[origin= c]{90}{\textit{inpainting}}} 
        & MVSS${^{*}}$
        & .215 & .150 & .100 & .064 & .136 & .083 & .073 & .045 \\
        & RIML${^{*}}$
        & .102 & .070 & .033 & .021 & .003 & .001 & .012 & .007 \\
        & DRAW-MVSS 
       & .829 & .753 & .676 & .574 & .115 & .076 & .513 & .407 \\
        & DRAW-RIML & \textbf{.949} & \textbf{.918} & \textbf{.889} & \textbf{.836} & \textbf{.406} & \textbf{.326} & \textbf{.849} & \textbf{.784}   \\
        & DRAW-HRNet 
        & .934 & .894 & .867 & .811 & .360 & .284 & .761 & .691 \\
        \hline
	\end{tabular}
\end{table}
}
\begin{table}
\footnotesize
\renewcommand{\arraystretch}{1}
    \caption{\textbf{Generalizability to untrained ISP pipelines or datasets}. ${\mathcal{P}}$ and ${\mathcal{D}}$ are trained on RAISE.}
    	\label{table_transfer}
    \centering
	\begin{tabular}{c|c|c|ccccc}
		\hline
		\multicolumn{2}{c|}{Test Item} & {Forgery} & 
		NoAtk &
		Rescale & JPEG70 & M-Blur & G-Blur\\
        \hline
        \multirow{6}{*}{\rotatebox[origin=c]{90}{\textit{ISP}}} & \multirow{3}{*}{OpenISP}
        & Spli. & .929 & .910 & .837 & .933 & .620 \\
        & & Copy. & .941 & .919 & .843 & .941 & .880  \\
        & & Inpa. & .850 & .820 & .451 & .765 & .756 \\
		\cline{2-8}
		& \multirow{3}{*}{Restormer}
        & Spli. & .946 & .936 & .863 & .941 & .648 \\
        & & Copy. & .961 & .947 & .871 & .948 & .904 \\
        & & Inpa. & .906 & .833 & .487 & .789 & .759 \\
		\hline

        \multirow{6}{*}{\rotatebox[origin=c]{90}{\textit{Dataset}}} & \multirow{3}{*}{Canon}
        & Spli.& .936 & .925 & .845 & .931 & .596 \\
        & & Copy. & .957 & .930 & .859 & .946 & .881  \\
        & & Inpa. & .805 & .732 & .486 & .710 & .706   \\
		\cline{2-8}
		& \multirow{3}{*}{SIDD}
        & Spli. & .928 & .909 & .832 & .911 & .574  \\
        & & Copy. & .967 & .965 & .891 & .954 & .880 \\
        & & Inpa. & .686 & .628 & .400 & .574 & .554\\
		\hline
	\end{tabular}
\end{table}

\noindent\textbf{Generalizability.}
We conduct additional experiments where $\mathcal{P}$ trained on RAISE dataset is applied on different RAW datasets, i.e., Canon and SIDD, and untrained ISP pipelines, i.e., OpenISP and Restormer. 
Table~\ref{table_transfer} shows that raw protection can generalize to untrained cameras and ISP pipelines while preserving promising detection capacity. 

To justify the generalizability to lossy transmission, we randomly handcraft 150 manipulated images, upload them onto several famous OSNs and download them for detection.
Also, we test the performance against dual JPEG and salt \& pepper attack ($p=5\%$) which are untrained types for DRAW.
As shown in Table~\ref{table_comparison_hybrid}, DRAW can effectively resist lossy OSN transmission, and its protection remains valuable against unknown lossy operations.


\noindent\textbf{Computational Complexity.}
We compare the computational requirements of MPF-Net in Table~\ref{table_complexity} with SegNet~\cite{segnet}, ShuffleNet~\cite{ma2018shufflenet}, U-Net~\cite{Unet} and ENet~\cite{Enet}, which are famous lightweight models for image segmentation. 
MPF-Net requires lower computing resources, e.g,  only 20.9\% in memory cost and 0.95\% in parameters compared to the classical U-Net.
\subsection{Baseline Comparisons}
\label{section_baseline}
Previous techniques in proactive image forgery detection, e.g., tag retrieval~\cite{wang2021faketagger} or template matching~\cite{asnani2022proactive}, are not suitable for image manipulation localization.
Moreover, Ying et al.~\cite{ying2021image}'s method additionally considers image self-recovery, which inevitably includes much heavier protective signal.
Therefore, we alternatively build two baseline methods that respectively apply pure robust training using our proposed attack layer and apply RGB-domain protection.
In the tests, MVSS is employed as localization network. 
The quantitative comparison results are reported in Table~\ref{table_comparison_baseline}. Further details regarding the experimental settings for the two baseline methods are included \redmarker{in the supplement}. 
\setlength{\tabcolsep}{1.05mm}{
\begin{table}[!t]
\footnotesize
    \caption{\textbf{Generalizability to lossy transmission and untrained perturbations.} Dataset: RAISE.}
    	\label{table_comparison_hybrid}
    \centering
	\begin{tabular}{c|cc|cc|cc|cc|cc}
		\hline
		\multirow{2}{*}{Forgery} & 
		 \multicolumn{2}{c|}{S\&P} &
		 \multicolumn{2}{c|}{Dual JPEG} &
		 \multicolumn{2}{c|}{Facebook} &
		 \multicolumn{2}{c|}{Weibo} &
		 \multicolumn{2}{c}{WeChat}\\
        \cline{2-11}
        & F1 & IoU 
        & F1 & IoU 
        & F1 & IoU
        & F1 & IoU
        & F1 & IoU\\
        \hline
        splicing
        & .839 & .855 & .657 & .683 & .917 & .920 & .902 & .897 & .763 & .728\\
        copymove
        & .854 & .850 & .692 & .729 & .905 & .910 & .859 & .870 & .637 & .688\\
        inpainting
        & .687 & .711 & .377 & .423 & .665 & .598 & .623 & .577 & .410 & .355\\
		\hline
  
	\end{tabular}
\end{table}
}
\setlength{\tabcolsep}{1.05mm}{
\begin{table}[!t]
\footnotesize
\renewcommand{\arraystretch}{1}
	\caption{\textbf{Comparison of computational cost} among lightweight image-to-image-translation or segmentation networks.}
	\centering
	\begin{tabular}{c|c|c|c|c|c}
		\hline
	    & SegNet~\cite{segnet} & ShuffleNet~\cite{ma2018shufflenet}  & U-Net~\cite{Unet} & ENet~\cite{Enet} & MPF-Net   \\
        \hline
        Params & 29.5M & 0.94M & 26.35M & 0.36M & 0.25M \\
        FLOPS & 0.56T & 22.9G & 0.22T & 2.34G & 7.39G \\
        Memory & 465MB & 390MB & 767MB & 46MB & 160MB\\
		\hline
	\end{tabular}
	\label{table_complexity}
\end{table}
}

\noindent\textbf{RAW Protection vs Pure Robust Training.}
Our proposed robust training mechanism reflected in the attack layer is different from that proposed in RIML. 
\ormarker{Specifically, }we render the unprotected RAW files $\mathbf{R}$ using $\mathcal{S}$, which are then attacked by $\mathcal{A}$.
We see that the introduction of robust training can help boost the performance of MVSS.
However, the overall performance is still worse than further applying RAW protection to aid localization. 
In severe degrading cases such as blurring, the performance gap between RAW protection and robust training without protection regarding F1 score is more than ten percent.

\noindent\textbf{RAW protection vs RGB protection.}
For fair comparison, we regulate that the overall PSNR on RGB images before and after RGB protection should be above 40 dB, in line with the criterion in Table~\ref{table_different_resolution_protected}. 
We conduct qualitative experiment in Fig~\ref{image_comparison} to evaluate the effectiveness of image protection. 
According to the experimental results, RGB protection cannot aid robust manipulation localization if the magnitude of RGB modification is restricted.
We also grayscale the augmented injected signal for better visualization and found
that signal injected by RAW protection is more adaptive in magnitude to the image contents. 
One possible reason is that the densely-predicting task requires hiding more information than binary image forgery classification task, making it struggle to maintain high fidelity of the original image.
In comparison, RAW protection can adaptively introduce protection with the help of content-related procedures, e.g., demosaicing and noise reduction, within the subsequent ISP algorithms that suppress unwanted artifacts and biases.
Theoretically, RAW data modification enjoys a much larger search space that allows transformations from the original image into another image with high density upon sampling. 
\setlength{\tabcolsep}{1.1mm}{
\begin{table}
\footnotesize
\renewcommand{\arraystretch}{1}
\caption{\textbf{Comparison with baseline methods on RAISE.}} We verify the importance of RAW protection by comparing the results with those of pure robust training using $\mathcal{A}$ and direct RGB protection. $\mathcal{P}^-$: using $\mathcal{P}$ for RGB protection. $\mathcal{D}$: MVSS$^*$
    	\label{table_comparison_baseline}
    \centering
	\begin{tabular}{c|cccc|cc|cc|cc|cc}
		\hline
		 \multirow{2}{*}{} &
		\multicolumn{4}{c|}{Used Modules} & 
		 \multicolumn{2}{c}{Rescaling} &
		 \multicolumn{2}{c}{JPEG70} &
            \multicolumn{2}{c}{Med. Blur} &  \multicolumn{2}{c}{GBlur} \\
        \cline{2-13}
        & ${\mathcal{P}}$ & ${\mathcal{P}^-}$ & ${\mathcal{A}}$ & ${\mathcal{D}}$
        & F1 & IoU 
        & F1 & IoU 
        & F1 & IoU 
        & F1 & IoU \\
        \hline
        \multirow{4}{*}{\rotatebox[origin=c]{90}{\textit{splicing}}} 
        & & & & \checkmark
        & .609 & .470 & .565 & .415 & .695 & .561 & .211 & .138  \\
        & & & \checkmark& \checkmark
       & \textbf{.668} & \textbf{.534} & .725 & .590 & .762 & .635 & .303 & .207  \\
        & & \checkmark& \checkmark& \checkmark
      & .358 & .253 & .438 & .317 & .487 & .361 & .149 & .097  \\
        & \checkmark & & \checkmark& \checkmark
        & .636 & .514 & \textbf{.789} & \textbf{.680} & \textbf{.770} & \textbf{.658} & \textbf{.419} & \textbf{.301}  \\
		\hline
		\multirow{4}{*}{\rotatebox[origin= c]{90}{\textit{copy-moving}}} 
        & & & & \checkmark
        & .636 & .544 & .471 & .366 & .731 & .640 & .336 & .258   \\
        & & & \checkmark& \checkmark
        & \textbf{.859} & \textbf{.816} & .648 & .582 & .782 & .728 & .528 & .456   \\
        & &\checkmark & \checkmark& \checkmark
         & .490 & .412 & .467 & .382 & .626 & .548 & .268 & .208   \\
        & \checkmark& & \checkmark& \checkmark
        & .836 & .780 & \textbf{.767} & \textbf{.706} & \textbf{.851} & \textbf{.803} & \textbf{.657} & \textbf{.582} \\
		\hline
		\multirow{4}{*}{\rotatebox[origin= c]{90}{\textit{inpainting}}} 
        & & & & \checkmark
        & .062 & .039 & .097 & .058 & .050 & .030 & .043 & .026 \\
        & & & \checkmark& \checkmark
        & .605 & .494 & .231 & .159 & .398 & .297 & .342 & .249 \\
        & & \checkmark& \checkmark& \checkmark
         & .387 & .291 & .480 & .371 & .381 & .288 & .374 & .279   \\
        & \checkmark& & \checkmark& \checkmark
        & \textbf{.682} & \textbf{.588} & \textbf{.536} & \textbf{.434} & \textbf{.595} & \textbf{.497} & \textbf{.561} & \textbf{.463}  \\
        \hline
	\end{tabular}
\end{table}
}
\subsection{Ablation Studies}
\label{section_ablation}
\begin{figure}[!t]
    \centering
    \includegraphics[width=0.48\textwidth]{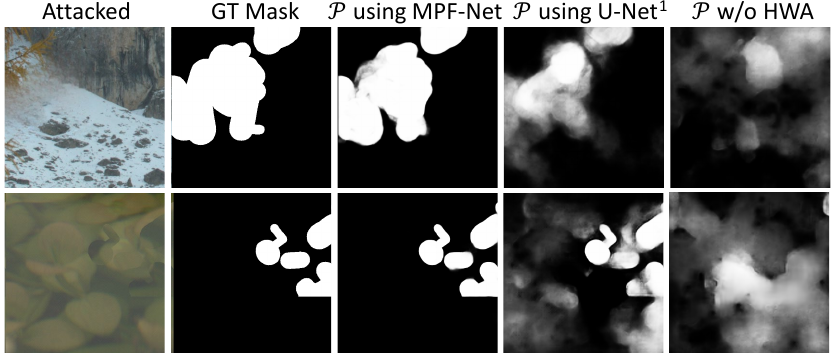}  \caption{\textbf{Examples of ablation studies of DRAW.} We observe that either replacing MPF-Net with U-Net using \textit{DSConv} or removing HFC module results in decreased performance. Upper: inpainting + JPEG80. Lower: copy-moving + median-blur.}
    \label{image_ablation}
\end{figure}

Table~\ref{table_ablation} and Fig.~\ref{image_ablation} respectively show the quantitative and qualitative results of ablation studies. 
In each test, we regulate that the averaged PSNR between $\mathbf{I}$ and $\hat{\mathbf{I}}$, with ISP pipelines evenly applied, should be within the range of 41-43 dB, to ensure imperceptible image protection.

\noindent\textbf{Substituting the architecture of $\mathcal{P}$.}
We first test if using U-Net with a similar amount of parameters or ENet~\cite{Enet} as $\mathcal{P}$ can achieve similar performance on splicing detection.
First, though ENet contains similar amount of parameters compared to MPF-Net, the performance of image manipulation localization using ENet as $\mathcal{P}$ is not satisfactory.
Second, though U-Net with $\textit{DSConv}$ provides much better result, because the channel numbers within each layer are restricted within 48 to save computational complexity, the performance is still worse than our benchmark.
\begin{table}[!t]
\footnotesize
\renewcommand{\arraystretch}{1}
\centering
  \caption{\textbf{Ablation study on DRAW on Nikon using splicing attack.} $^1$: replacing \textit{Conv} layers with \textit{DSConv}.}
\begin{tabular}{cccc}
\hline
\multirow{2}{*}{Test} & \multicolumn{3}{c}{F1} \\
  \cline{2-4}  & NoAtk  & JPEG70 & Mblur \\

\hline
$\mathcal{P}$ using U-Net$^1$  & .877  & .769  & .535   \\
$\mathcal{P}$ using ENet   & .324  & .137   & .092   \\
\hline
MPF-Net w/o HFC & .844  & .710 & .602\\
MPF-Net w/o DT-CWT & .852 & .751 & .626\\
MPF-Net w/o PFF & .827 & .712 & .667\\
\hline
w/o diff from real attack & .842  & .566 & .502\\
using only one ISP Surrogate & .648  & .455 & .267\\
w/o Image Distortion Module & .929  & .245  & .116\\
w/o Color Ajustment Module & .814  & .759  & .641\\
\hline
Full implementation of \textbf{DRAW}  & .929  & \textbf{.838} & \textbf{.696} \\
\hline
\end{tabular}
\label{table_ablation}
\end{table}

\noindent\textbf{Impact of components in MPF-Net.}
The most noticeable difference between MPFNet with previous U-shaped networks is that feature disentanglement can be better ensured even with fewer parameters.
To verify this, we respectively replace the HFC layer and PFF layer with typical alternatives, i.e., vanilla convolution and channel-wise concatenation.
The performances are nearly 5-10 points weaker compared to the MPF-Net setup.
First, DT-CWT is a shift-invariant wavelet transform that comes with limited redundancy.
Second, partial feature fusion and partial connection are more flexible.
The design explicitly keeps some of the features extracted from the current level and directly feeds them into the subsequent block. Therefore, for different levels, the input features will be different, which encourages feature disentanglement.

\noindent\textbf{Impact of pipeline design.}
We also tested the setting of not using the image distortion module or color adjustment module in the pipeline during training.
The result is as expected that the scheme will therefore lack generalizability in overall robustness due to the fact that there are not enough random processes that can simulate the real-world situation.
Besides, not introducing the difference between the real-world and simulated attacks or using only one ISP surrogate model will also impair the overall performance.


\section{Conclusions}
We present DRAW that adds imperceptible protective signal to the RAW data against image manipulation. The protection can be transferred into RGB images and resist lossy operations.
Extensive experiments on typical RAW datasets prove the effectiveness of DRAW.

\medskip
\noindent\textbf{Acknowledgment}. This work was supported by National Natural Science Foundation of China under Grant U20B2051, U1936214, U22B2047, 62072114 and U20A20178. 

{\small
\bibliographystyle{ieee_fullname}
\bibliography{iccv_final}
}

\end{document}


\title{Supplementary Material of ``DRAW: Defending Camera-shooted RAW against Image Manipulation"}
\maketitle

\input{command}
In this supplementary material, we provide the details of the hybrid attack layer, the baseline designs and the experimental settings.
Also, more experimental results on the imperceptibility of RAW protection and the performance of robust image manipulation detection are presented.
Finally, we extend our RAW protection method by applying it to the existing RGB images from two famous image manipulation detection datasets, i.e., CASIA~\cite{CASIA} and Defacto~\cite{Defacto}, with the help of RGB2RAW inversion.
The code for implementing DRAW is included in the attached file, where the ``README.md" file shows detailed procedures how to generate protected images, how to conduct automatic image manipulation and how to determine the forged areas. Due to file size limit, we are only able to include the checkpoint of the protection model, while those of the localization networks are too large to upload, e.g., 252MB for HRNet and 163MB for MVSS.

\section{Details of the Hybrid Attack Layer.}
\noindent\textbf{Manipulation Mask Generation.}
Real-world image tampering may be oriented on regions of interest within a targeted image. 
However, code-driven realistic image manipulation can be expensive and time-consuming. 
We study the natural distribution of tampered areas by observing the binary masks in CASIA dataset~\cite{CASIA}.
The location of forgery within an image roughly follows a uniform distribution except for corners, and for most manipulated images, the total area of forged contents is within the range of 5\%-30\%. 
For simplification, we assume that the location of forgery within an image during training roughly follows a uniform distribution and the accumulated manipulated squared area is within the range of $[0,0.3]$.

We apply free-form mask generation~\cite{yu2019free} to arbitrarily select areas within $\hat{\mathbf{I}}$ according to a binary mask $\mathbf{M}$. 
\begin{equation}
\label{eqn_manipulate}
\hat{\mathbf{I}}^{\emph{t}}=\hat{\mathbf{I}}\cdot(1-\mathbf{M})+\mathbf{R}\cdot\mathbf{M},
\end{equation}
where $\mathbf{R}$ is the source of manipulation.

\noindent\textbf{Image Manipulation Simulation.}
For image manipulation, we simulate the most common types of tampering, which include \textit{copy-move}, \textit{splicing} and \textit{inpainting}.
The simulation of different kinds of attacks can be reflected by the composition of $\mathbf{R}$ in Eq.~(\ref{eqn_manipulate}).
For \textit{copy-move}, we let $\mathbf{R}$ in Eq.~(\ref{eqn_manipulate}) as a spatially-shifted version of $\hat{\mathbf{I}}^{\emph{t}}$. 
For \textit{splicing}, we use another random RGB image as $\mathbf{R}$. 
However, we find that this setting of attack will encourage the network to widen the distribution gap between $\hat{\mathbf{I}}$ and natural RGB images to better distinguish each other, thus greatly decreasing the overall image quality.
To address this, we also apply an enhanced \textit{splicing} attack named \textit{coincident-splicing} that ``coincidentally" use another protected RGB $\hat{\mathbf{I}}'$ as $\mathbf{R}$.   
For \textit{inpainting}, we use the open-source model from LAMA~\cite{LAMA} and ZITS~\cite{ZITS} to generate the inpainted result as $\mathbf{R}$.
We iteratively and evenly perform the above \textit{three} types of attacks for balanced training. 

\noindent\textbf{Image Distortion Simulation.}
Similar to HiDDeN~\cite{zhu2018deep}, we simulate typical image lossy post-processing operations to enhance the robustness of the proposed method.
The involved attacks include the following: 
(1) \textit{rescaling}, which resizes the image by an arbitrarily resizing rate $r\in[50\%,150\%]$, 
(2) \textit{median blurring}, which blurs the image using median filter whose kernel size $k$ is arbitrarily selected from $[3,5]$, 
(3) \textit{Additive White Gaussian Noise} (AWGN), which adds Gaussian noise evenly on the image, where the standard value s ranges from zero to one, 
(4) Gaussian blurring, which is similar to the median blurring but the kernel is different,
(5) \textit{JPEG compression}, which compresses the image using the popular Diff-JPEG~\cite{shin2017jpeg} with tunable JPEG quality factors.

\noindent\textbf{Color Adjustment Simulation.}
Most users prefer manually adjusting the brightness or contrast after RAW files are automatically rendered into RGBs.
Therefore, we also simulate typical color adjustment operations to mitigate their impacts on the performance of our method.
The involved attacks include the following:
(1) \textit{Hue adjustment}: the image hue is adjusted by converting the image to HSV and cyclically shifting the intensities in the hue channel. The image is then converted back to the original image mode.
The hue factor is set within the range of $[-0.05.0.05]$.
(2) \textit{contrast enhancement}: we adjust the contrast of an image, where the contrast factor is set within the range of $[0.7,1.5]$.
(3) \textit{saturation adjustment}: we adjust the color saturation of the image, where the factor is set within the range of $[0.7,1.5]$.
(4) \textit{brightness adjustment}: we adjust the brightness of the image, where the factor is set within the range of $[0.7,1.5]$.
The differentiable data augmentation functions applied during training are implemented by the APIs from the ``torchvision" package.

\noindent\textbf{Attack Generation during Testing.}
For quantitative evaluation for \textit{splicing}, we borrow the segmentation masks from MS-COCO~\cite{lin2014microsoft} dataset, crop out the corresponding objects and iteratively add them onto the protected images $\hat{\mathbf{I}}$ until the total manipulation rate exceeds 5\%. 
For \textit{copy-move}, we generate the attacked image under the same principle that was used during training.

\noindent\textbf{Real-world Attack Involvement.}
The real-world attacks are implemented by the APIs from the ``cv2" package, e.g., \textit{cv2.GaussianBlur} for Gaussian blurring and \textit{cv2.imencode} for JPEG compression.
These functions are performed on PIL images in ``ndarray" format, which requires that we transform the 32-bit float-typed tensors into 8-bit integer-based arrays.
Therefore, quantization attack is also automatically considered by the introduction of real-world attacks. 
In each iteration of the training stage, we perform the corresponding real-world attacks using the same setting from the simulated methods.

\section{Frequency Learning in Deep Networks}
\noindent\textbf{Existing Methodologies.}
Frequency-learning is an efficient way to reduce computation resource costs. For example, Xu et al.~\cite{xu2020learning} proposes a learning-based frequency selection method to identify trivial frequency components in the input images, which can be removed without performance loss. According to ~\cite{lee2021fnet}, self-attention layers can be replaced with simple Fourier transformations to spped up Transformer encoder architectures under limited accurary sacriface. FEDformer~\cite{zhou2022fedformer} exploits the sparse representation in Fourier transform to capture the global view of time series. 
In addition, frequency-domain information has shown great potential in revealing subtle differences between real and fake images, such as in face forgery detection tasks, where it can help detect generated faces~\cite{dzanic2020fourier, chandrasegaran2021closer, wang2020cnn} or synthesized images~\cite{qian2020thinking, li2021frequency, liu2021spatial} based on face-swapping techniques.

However, the above-listed work only replaces interpolation with DWT or DCT, which still requires heavy computation. In order to design a new lightweight network with frequency learning, we must effectively combine the advantages of wavelet transform and CNN architecture.

\begin{figure}[!t]
    \centering
    \includegraphics[width=0.5\textwidth]{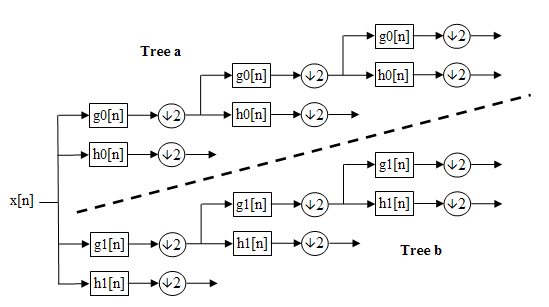}
    \caption{\textbf{Illustration of DT-CWT,} which is a two-dimensional transform that decomposes an image into six frequency subbands at each level of the transform.}
    \label{image_DTCWT}
\end{figure}
\noindent\textbf{DT-CWT Transformation.}
The Dual-Tree Complex Wavelet Transform (DT-CWT) is a type of wavelet transform used in signal and image processing. 
It was introduced by Kingsbury~\cite{DTCWT} and is an extension of the discrete wavelet transform (DWT) that uses complex wavelets.
The DT-CWT is a two-dimensional transform that decomposes an image into six frequency subbands at each level of the transform. These subbands are formed by filtering the image with two sets of filters, one for the real part of the wavelet and the other for the imaginary part. 
The filters are designed to have good directional selectivity and to be approximately shift-invariant.

The DT-CWT has several advantages over other wavelet transforms, e.g., the Haar Wavelet Transform~\cite{stankovic2003haar} and the Daubechies Wavelet Transform~\cite{vonesch2007generalized}. 
1) Directionality. The DT-CWT is designed to be directional, meaning it can capture edges and other features that have a preferred orientation. This is particularly useful in image processing tasks, where many objects have a characteristic orientation.
2) Phase information. The DT-CWT provides complex coefficients, which contain both magnitude and phase information. This phase information can be useful in certain applications, such as texture analysis, where the phase information can help distinguish between different types of textures.
3) Redundancy. The DT-CWT is redundant, meaning that some of the information is duplicated across different subbands. This redundancy can be useful in denoising or other signal processing tasks, where the redundant information can help to improve the robustness of the algorithm.

\begin{figure*}[!t]
    \centering
    \includegraphics[width=1.0\textwidth]{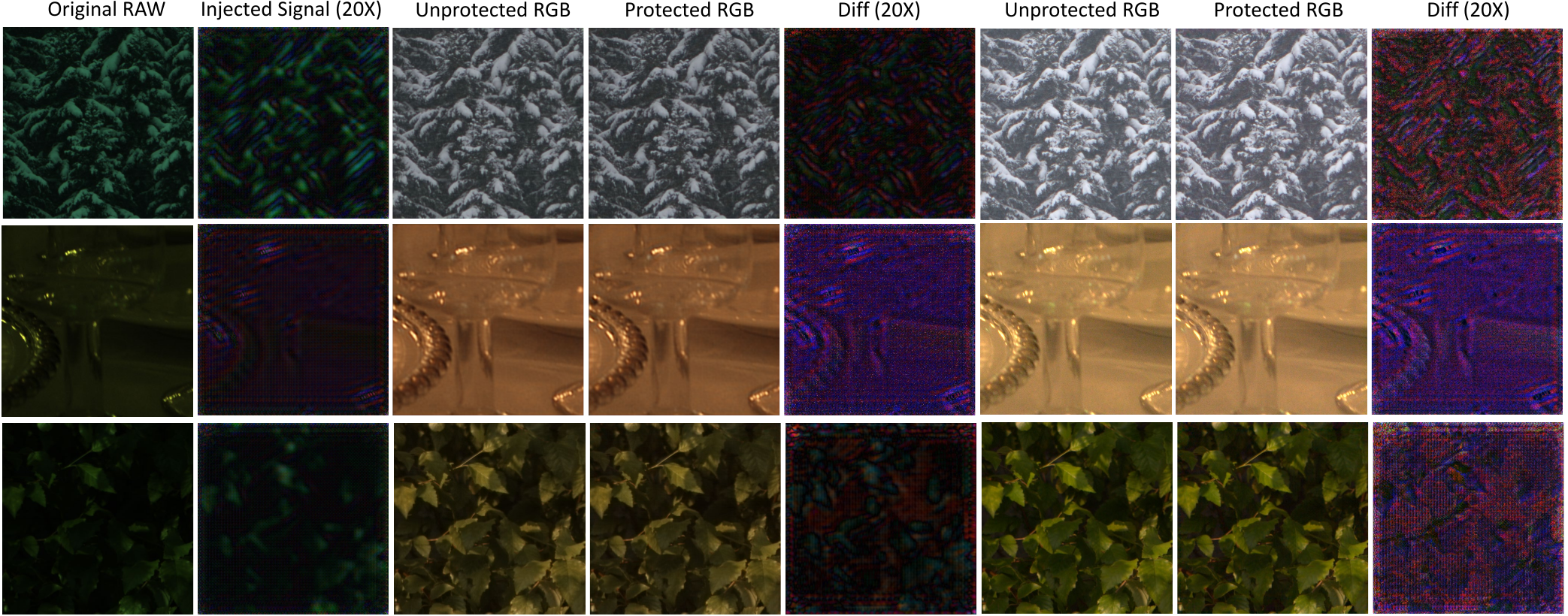}
    \caption{\textbf{Examples of protected RAWs and the corresponding protected RGBs.} In each test, we apply two ISPs for rendering (upper: InvISP / TradISP, middle: LibRAW / TradISP, lower: CycleISP / LibRAW).  The RAW images are visualized through bilinear demosaicing.}
    \label{image_protection_isp}
\end{figure*}

The DT-CWT has been successfully applied to various computer vision tasks, including image denoising~\cite{hill2012undecimated}, image super-resolution~\cite{izadpanahi2013motion}, and object detection~\cite{sun2006face,sengar2020moving}. For example, in object detection, the DT-CWT can be used to extract features that are both scale and orientation invariant, which can improve the accuracy of the detector. In image super-resolution, the DT-CWT can be used to extract high-frequency information that is lost during image downsampling, which can then be used to reconstruct a higher-resolution image.
With the development of modern CNN networks, researchers prefer learning end-to-end feature extractors in favor of pre-designed filters, which possibly results in a downgraded role of wavelet transform played in computer vision tasks. However, compared to cascaded learnable convolutional layers, DT-CWT transform still contain several advantages as follows.

\noindent\textbf{Bringing DT-CWT into CNNs.}
Introducing DT-CWT into CNNs can have several advantages for our task and beyond. First, the DT-CWT is robust to noise in image data, as it can extract features at multiple scales and orientations. This can help improve the performance of CNNs on modifying the higher-frequency details.
Second, the DT-CWT can extract rich features that are both scale and orientation invariant, which can improve the discriminative power of CNNs. This can be particularly useful in content-aware protective signal embedding.
Thirdly, the DT-CWT can reduce the complexity of CNNs by reducing the number of filters required in the initial layers. It also provides both magnitude and phase information, which can be used to visualize and interpret the learned features.

In MPF-Net, we combine the benefit of DT-CWT with Fourier frequency learning, where FFT can mitigate the issue of focusing too much on local patterns. 
Besides, global information aggregation and lower computational complexity is achieved by the proposed HFC and PFF mechanisms.




\section{More Experimental Results}

\begin{figure}[!t]
    \centering
    \includegraphics[width=0.48\textwidth]{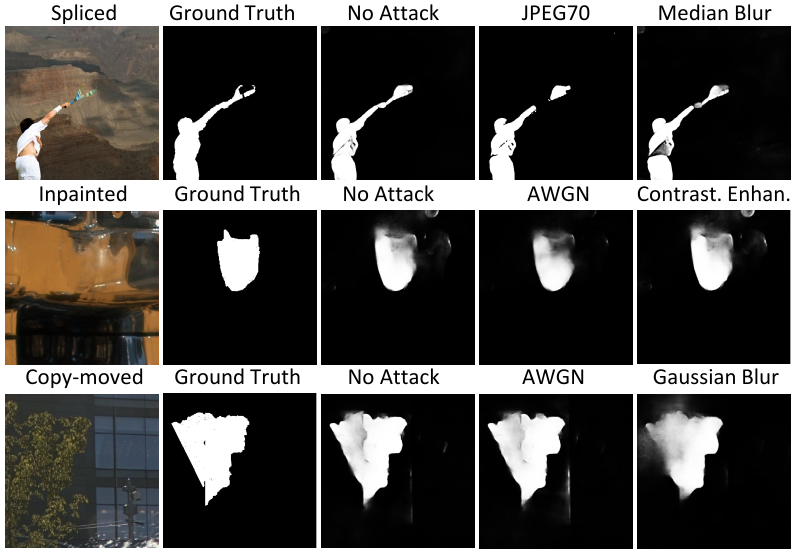}
    \caption{\textbf{Exampled image manipulation detection results of DRAW-HRNet.} Multiple image lossy operations are involved.}
    \label{image_detection_hrnet}
\end{figure}

\begin{figure*}[!t]
    \centering
    \includegraphics[width=1.0\textwidth]{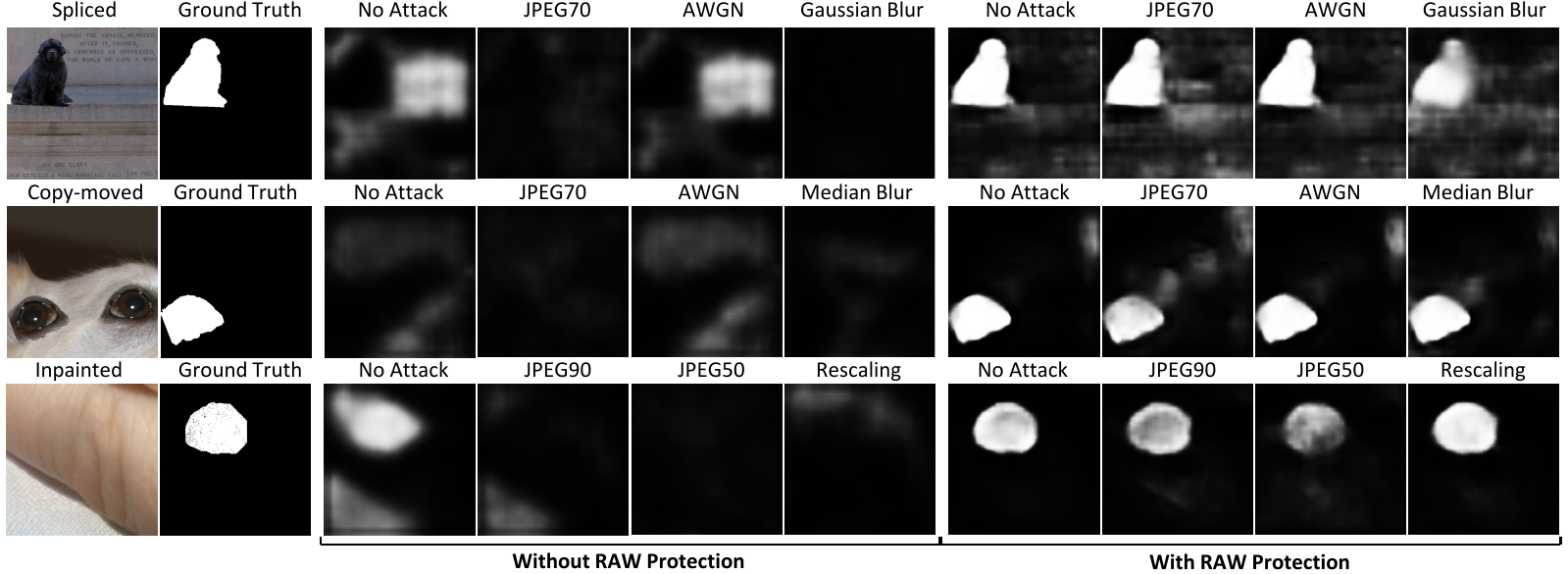}
    \caption{\textbf{Example of performance gain on MVSS with DRAW.} The protection helps the detector locate the forged area despite the presence of lossy image operations.}
    \label{image_improve_MVSS}
\end{figure*}

\begin{figure*}[!t]
    \centering
    \includegraphics[width=1.0\textwidth]{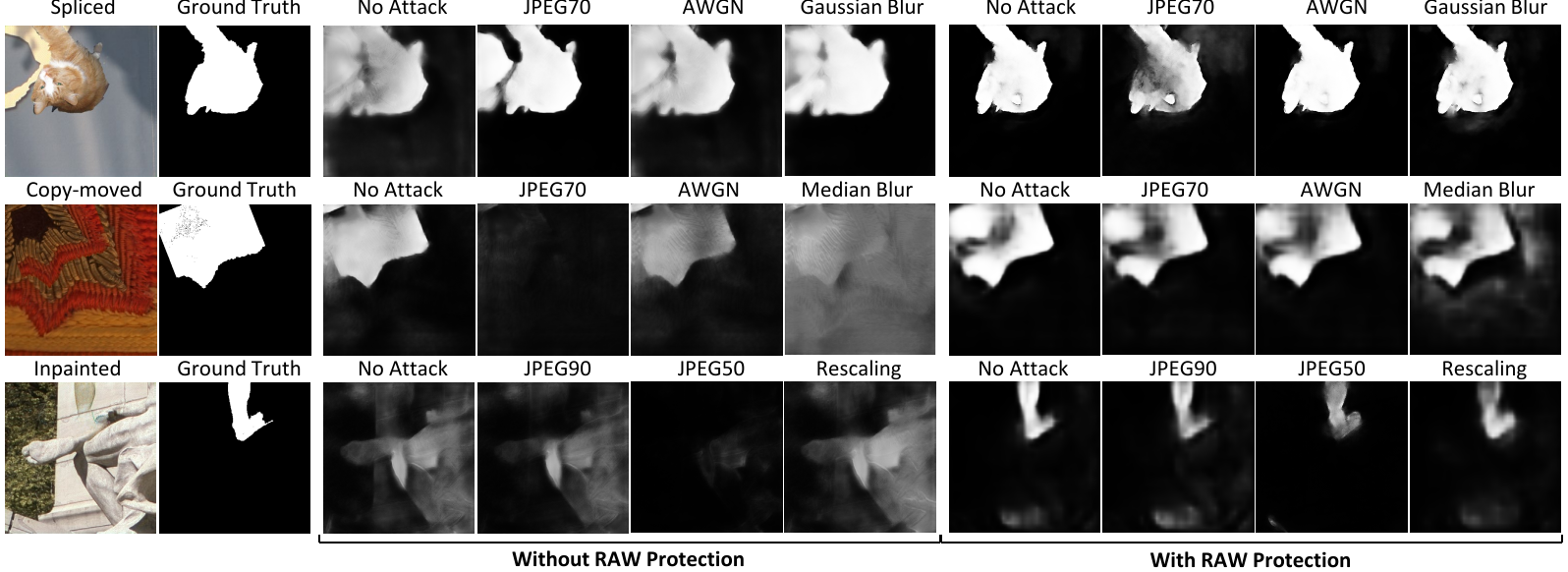}
    \caption{\textbf{Example of performance gain on RIML with DRAW.} The protection helps the detector locate the forged area despite the presence of lossy image operations. }
    \label{image_improve_RIML}
\end{figure*}

\setlength{\tabcolsep}{1.05mm}{
\begin{table}[!t]
\footnotesize
    \caption{\textbf{Average performance against color adjustment attacks and hybrid attacks on RAISE dataset.} The detector can successfully locate the forged areas in most cases.}
    	\label{table_comparison_hybrid}
    \centering
	\begin{tabular}{c|ccc|ccc|ccc}
		\hline
		\multirow{2}{*}{Attack} & 
		 \multicolumn{3}{c|}{splicing} &
		 \multicolumn{3}{c|}{copy-move} &
		 \multicolumn{3}{c}{inpainting} \\
        \cline{2-10}
        & Rec. & F1 & IoU 
        & Rec. & F1 & IoU 
        & Rec. & F1 & IoU \\
        \hline
        Hue Adjust.
        & .938 & .949 & .905 & .973 & .974 & .962 & .779 & .794 & .736 \\
        Contra. Enhan.
        & .935 & .945 & .900 & .971 & .969 & .958 & .773 & .783 & .726  \\
        Satur. Adjust.
        & .937 & .948 & .904 & .969 & .968 & .958 & .784 & .795 & .738 \\
        Bright. Adjust.
        & .936 & .947 & .903 & .960 & .960 & .948 & .771 & .782 & .725  \\
		\hline
		JPEG70+Hue.
        & .900 & .855 & .769 & .906 & .872 & .824 & .489 & .508 & .396  \\
        G-Blur+Contra.
        & .553 & .637 & .520 & .895 & .902 & .866 & .755 & .774 & .692 \\
        M-Blur+Satur.
        & .927 & .939 & .890 & .960 & .956 & .938 & .821 & .832 & .754  \\
        AWGN+Bright.
        & .930 & .935 & .885 & .952 & .944 & .925 & .842 & .842 & .778 \\
		\hline
  
	\end{tabular}
\end{table}
}

Fig.~\ref{image_protection_isp} and Fig.~\ref{image_detection_hrnet} show more experimental results on the imperceptibility of RAW protection and the performance of robust image manipulation detection after applying for RAW protection.
From the results, we see that the injected signal is weak and the generated protected RGB images are not affected in their overall visual quality.
While previous forgery detection methods strive hard to find traces for unveiling manipulation with the presence of loss image operations, we could successfully locate the tampered area on the forged protected image. 
In Table~\ref{table_comparison_hybrid}, we further conduct color adjustment attacks and hybrid attacks on the protected images and let the networks detect the forged areas. The experiments show that our network can also detect forged areas, which proves the effectiveness of our scheme.

Fig.~\ref{image_improve_MVSS} and Fig.~\ref{image_improve_RIML} respectively show some examples of performance gain on MVSS~\cite{dong2021mvss} and RIML~\cite{RIML} with DRAW. 
The protection helps the two detectors locate the forged area despite the presence of lossy image operations.

\section{Details of the Baseline Methods.}
\begin{figure}[!t]
    \centering
    \includegraphics[width=0.48\textwidth]{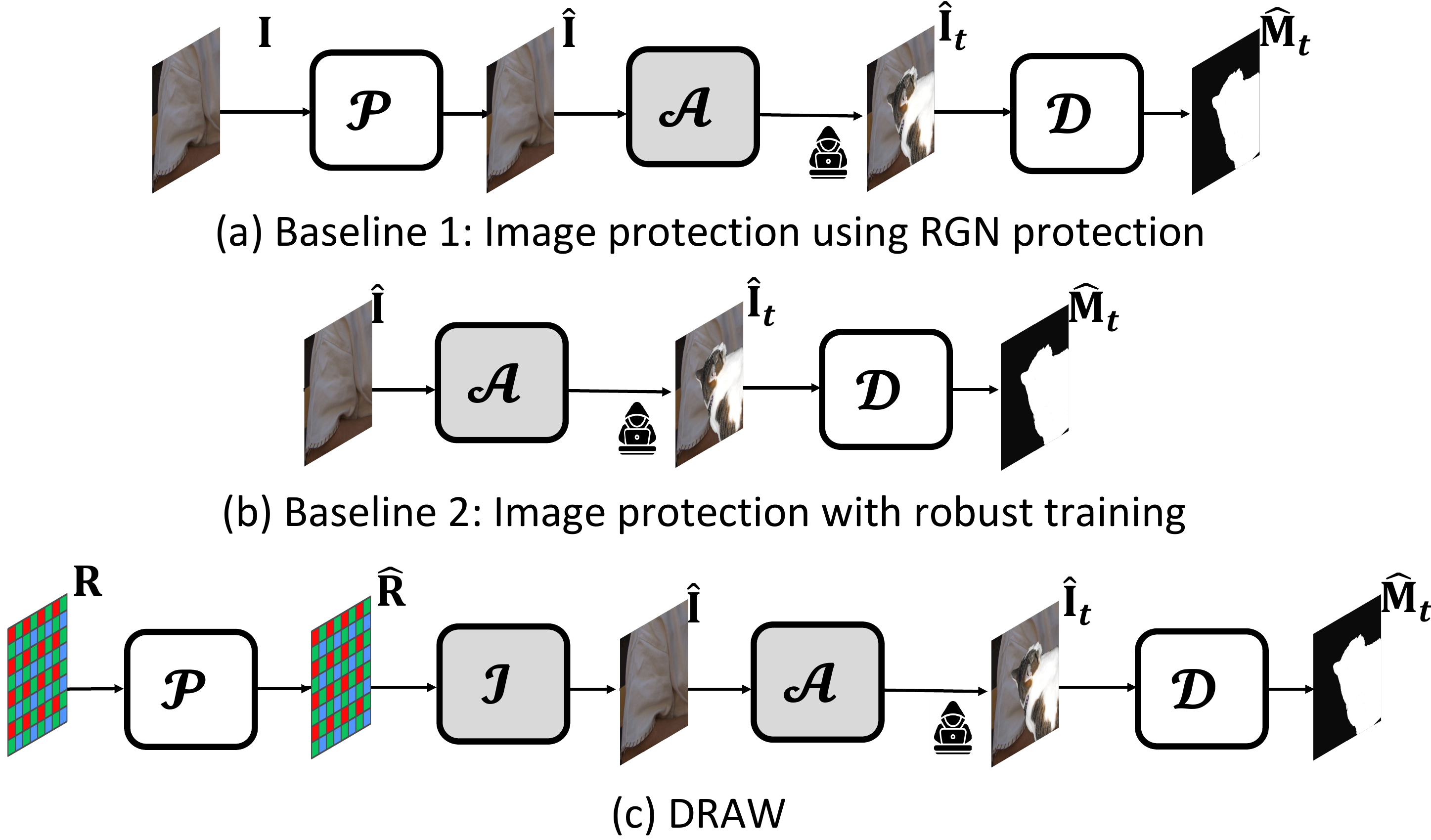}
    \caption{\textbf{Pipeline comparison between two baselines and DRAW.} We study the impact of RAW protection on the overall performance by making two substitutions in the network design.}
    \label{fig_baseline_illustrate}
\end{figure}

Fig.~\ref{fig_baseline_illustrate} illustrates the pipeline overview of the two baseline methods, namely, image forgery detection with pure robust training and image forgery detection using RGB protection.
Detailed settings are specified as follows.

\noindent\textbf{RAW Protection vs Pure Robust Training.}
We validate the impact of RAW protection on the performance of DRAW by first removing the RAW protection stage.
The corresponding fidelity terms are also removed.
In this case, no camera imaging pipeline is considered and the training technique of hybrid attacking layer involvement is solely responsible for improving the robustness of image manipulation localization, which is close to RIML.
According to the experiments, the baseline can indeed noticeably boost the performance of manipulation detection under lossy image operations, but the overall accuracy is worse than that of DRAW.
The reason is that finding a universally-existing trace to unveil manipulation is difficult in the real world, unlike injecting an outer-sourced signal for later retrieval.

\begin{figure}[!t]
    \centering
    \includegraphics[width=0.4\textwidth]{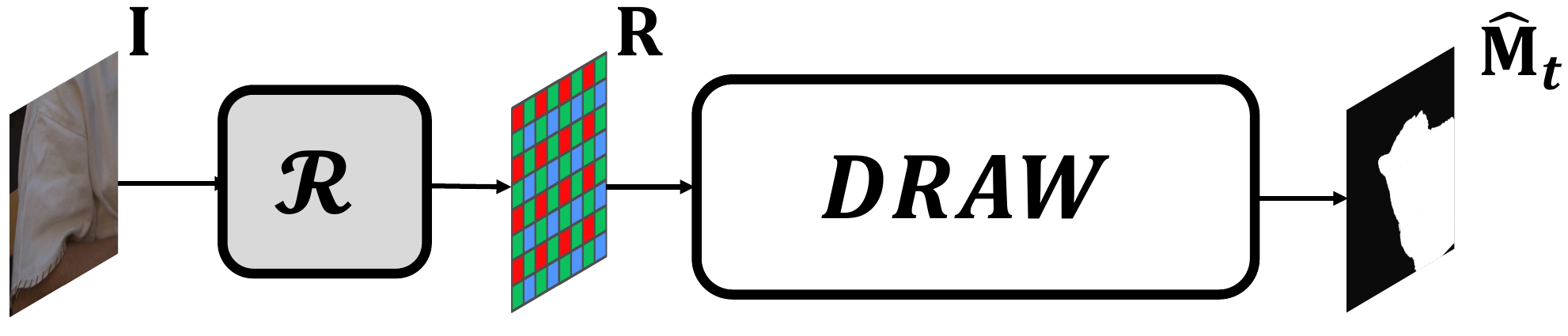}
    \caption{\textbf{Pipeline of RAW protection on the existing RGB images.} A revertion network is trained to revert an RGB image into approximated RAW data, which is then injected with protective signal to combat nefarious manipulations.}
    \label{fig_invert_illustrate}
\end{figure}

\begin{figure*}[!t]
    \centering
    \includegraphics[width=1.0\textwidth]{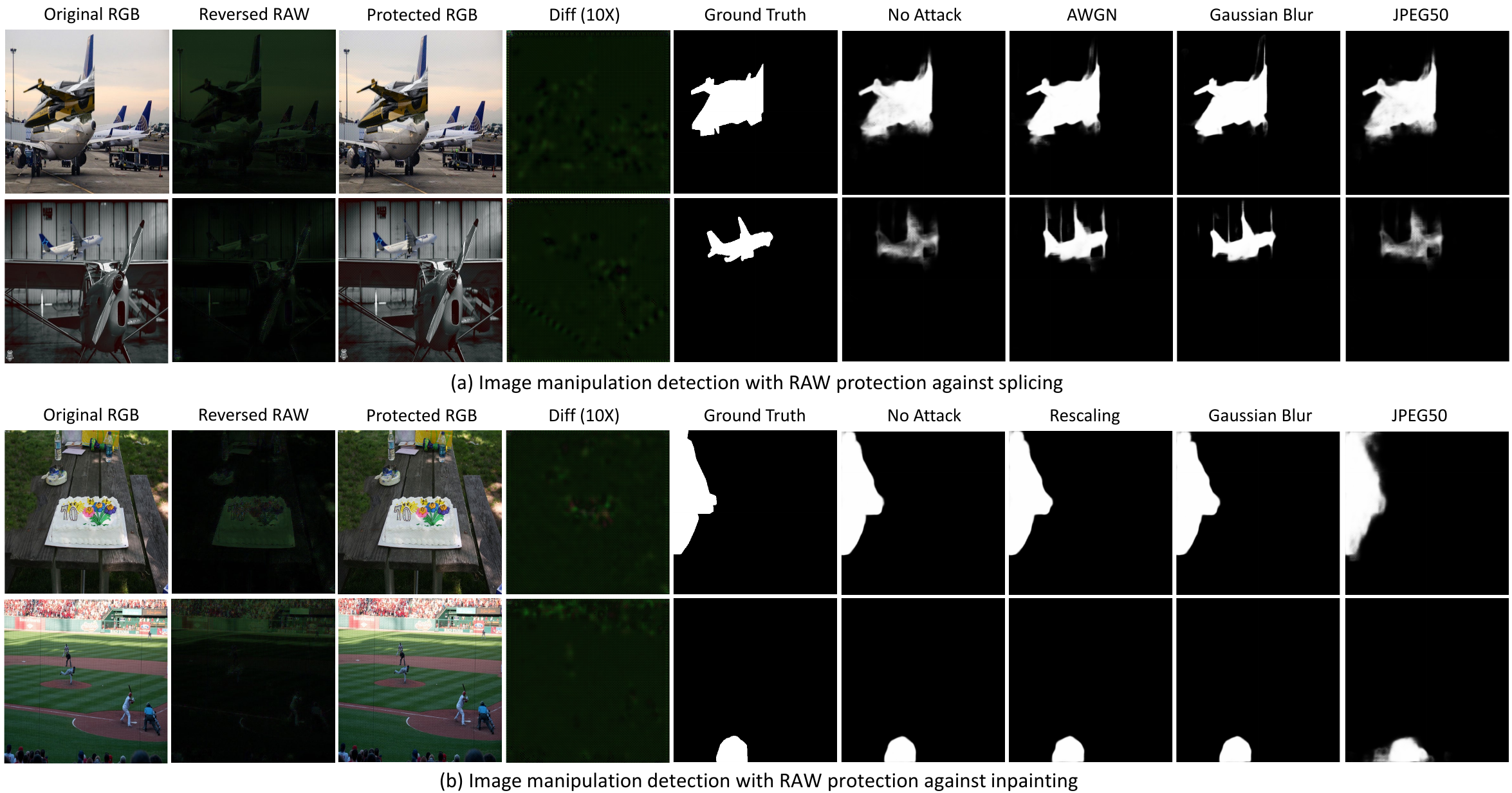}
    \caption{\textbf{Performance analysis of RAW protection on the existing RGB images.} We first revert the RGB images into RAW data using InvISP, and then perform RAW protection against image manipulation.}
    \label{fig_results_casia}
\end{figure*}

\noindent\textbf{RAW protection vs RGB protection.}
We compare RAW protection with RGB protection, in which we modify the original image for anti-manipulation protection. The ISP process is also ruled out in the pipeline.
The RAW protection term is therefore removed and the hyper-parameters are changed as $\beta=1,\gamma=0.01, \epsilon=0.005$.
Though the two schemes ideally can come up with the same solution where after image rendering, the protective signal embedded within RAW could be the same or close compared with that embedded directly within RGB, the experimental results show that successful RGB protection is more difficult compared to RAW protection.
The reason is that RAW data offers a much larger dynamic range compared to 8-bit RGB images, therefore offering way more possible solutions for effective image protection.

\section{Other Implementation Details}
We train all network-based ISP pipelines using RGB images rendered by the libraw library as supervision and these pre-trained ISPs will be frozen when training the RAW protection network.
We find that for different RAW datasets, the performances of cross-dataset RGB image rendering of ISP networks are not satisfactory.
Therefore, for each RAW dataset, we separately train their exclusive ISP networks.
In contrast, our protection network is transferable, and we train the network based on a single benchmark dataset, e.g., RAISE, and conduct experiments on other datasets on this model without further fine-tuning.

\setlength{\tabcolsep}{1.1mm}{
\begin{table}
\footnotesize
\renewcommand{\arraystretch}{1}
    \caption{\textbf{Average performance of RAW protection applied on the existing RGB images.} We revert images from typical manipulation datasets and perform RAW protection.}
    	\label{table_comparison_supp}
    \centering
    
	\begin{tabular}{c|cc|cc|cc|cc}
		\hline
		\multirow{2}{*}{Datasets} & 
		 \multicolumn{2}{c}{No attack} &
		 \multicolumn{2}{c}{Rescaling} &
            \multicolumn{2}{c}{AWGN} &  \multicolumn{2}{c}{GBlur} \\
        \cline{2-9}
        & F1 & IoU 
        & F1 & IoU 
        & F1 & IoU 
        & F1 & IoU \\
        \hline
        CASIA v2.0
        & .911 & .855 & .771 & .676 & .854 & .774 & .679 & .563  \\
        Defacto-splicing
        & .806 & .757 & .858 & .811 & .810 & .760 & .869 & .823  \\
        Defacto-inpainting
        & .869 & .812 & .902 & .846 & .864 & .807 & .920 & .862   \\
		\hline
	\end{tabular}
\end{table}
}
\section{Extending RAW Protection on the Existing RGB Images}

We further extend our method to include RAW protection into existing RGB images.
Considering that most RGB images do not have their corresponding RAW data, we train an RGB2RAW reversion network, denoted as $\mathcal{R}$, using InvISP~\cite{InvISP}, which is based on an invertible neural network~\cite{dinh2014nice}.
We train the reversion network based on the RAISE dataset. In the forward pass, the RAW data is transformed into the corresponding RGB image, and the RAW data is retrieved in the backward pass given the ground-truth RGB image.
After training this network, we feed it with the RGB images from CASIA and Defacto and generate their approximate RAW data.
Because the reverting network is trained on the RAISE dataset, we also use the ISP network trained on the RAISE dataset.
The pipeline is shown in Fig~\ref{fig_invert_illustrate}.

Note that CASIA and Defacto only provide forgery images and the manipulation masks, with no original clean images given.
Alternatively, after reverting RGBs into RAWs, we protect the entire image as a whole that includes the forged area, after image rendering, we alter the manipulated content using Eq.~(\ref{eqn_manipulate}), where the protective signal within the manipulated areas will be therefore lost.
We skip the testing on the \textit{copy-move} attack because we cannot obtain the original location of the source and therefore cannot introduce the shifted version of the protective signal on the forged area.
We use the default HRNet as the detector.

In Table~\ref{table_comparison_supp} and Fig.~\ref{fig_results_casia}, we respectively present the quantitative and qualitative experimental results of extending RAW protection on the images from CASIA and Defacto.
The averaged PSNR between the protected RGB images and their original versions is 42.15dB on CASIA and 41.43dB on Defacto, which shows pleasing visual results. 
Besides, the quantitative results on image manipulation detection show that even if we post-process the already forged images, the detector can successfully locate the forged areas.
The results show a promising direction of involving RAW protection in the study of image trust and authenticity where once the images are shot, they can be no longer easily manipulated at one's free will.

{\small
\bibliographystyle{ieee_fullname}
\bibliography{egbib}
}

%% file: command.tex
\definecolor{green}{rgb}{0, 0.5, 0}
\definecolor{orange}{rgb}{0.8, 0.6, 0.2}
\definecolor{orange2}{rgb}{1.0, 0.4, 0.5}
\definecolor{red}{rgb}{1.0, 0.0, 0.0}
\definecolor{teal}{rgb}{0.0, 0.4, 0.6}
\definecolor{purple}{rgb}{0.65,0,0.65}
\definecolor{saffron}{rgb}{0.95,0.75,0.2}
\definecolor{turquoise}{rgb}{0.0,0.5,0.5}
\definecolor{black}{rgb}{0.0, 0.0, 0.0}
\definecolor{gray}{rgb}{0.5, 0.5, 0.5}
\definecolor{brown}{rgb}{0.6, 0.3, 0.0}

\newcommand{\share}[1]{{\color{black}#1}}
\newcommand{\Rone}[1]{{\color{black}#1}}
\newcommand{\redmarker}[1]{{\color{black}#1}}
\newcommand{\greenmarker}[1]{{\color{black}#1}}
\newcommand{\Rtwo}[1]{{\color{black}#1}}
\newcommand{\Rthree}[1]{{\color{black}#1}}
\newcommand{\highlight}[1]{{\color{black}#1}}
\newcommand{\Rfour}[1]{{\color{black}#1}}
\newcommand{\ormarker}[1]{{\color{black}#1}}
